\documentclass{article}

     \PassOptionsToPackage{numbers, compress}{natbib}

\usepackage[preprint]{neurips_2022}

\usepackage[utf8]{inputenc} 
\usepackage[T1]{fontenc}    
\usepackage{hyperref}       
\usepackage{url}            
\usepackage{booktabs}       
\usepackage{amsfonts}       
\usepackage{nicefrac}       
\usepackage{microtype}      
\usepackage{xcolor}         

\hypersetup{colorlinks=true,linkcolor=red}
\usepackage{amsmath}
\usepackage{amssymb}
\usepackage{mathtools}
\usepackage{amsthm}
\usepackage{color, soul}

\theoremstyle{plain}
\newtheorem{theorem}{Theorem}[section]
\newtheorem{proposition}[theorem]{Proposition}
\newtheorem{lemma}[theorem]{Lemma}

\theoremstyle{definition}

\theoremstyle{remark}

\usepackage[]{enumitem}
\usepackage[]{listings}
\usepackage{caption}

\title{Two Instances of Interpretable Neural Network for Universal Approximations}

\author{%
  Erico Tjoa\thanks{Also affiliated to Alibaba Inc via Alibaba-NTU JRI} \\
  Interdisciplinary Graduate School\\
  Nanyang Technological University\\
  50 Nanyang Ave, 639798 \\
  \texttt{ericotjo001@e.ntu.edu.sg} \\
   \And
   Guan Cuntai \\
   School of Computer Science and Engineering \\
   Nanyang Technological University\\
   50 Nanyang Ave, 639798 \\
   \texttt{ctguan@ntu.edu.sg} \\
}

\begin{document}

\maketitle

\begin{abstract}
This paper proposes two bottom-up interpretable neural network (NN) constructions for universal approximation, namely Triangularly-constructed NN (TNN) and Semi-Quantized Activation NN (SQANN). Further notable properties are (1) resistance to catastrophic forgetting (2) existence of proof for arbitrarily high accuracies (3) the ability to identify samples that are out-of-distribution through interpretable activation ``fingerprints" .
\end{abstract}

\section{Introduction}
\label{section:intro}

Improving the interpretability of a deep neural network (deep NN or DNN) is important as it enables more transparent and responsible usage. However, its black-box nature remains a difficult challenge to the machine learning (ML) community. The challenge has given rise to the eXplainable artificial intelligence; various studies to understand how a DNN works have correspondingly emerged (see the following surveys \cite{BARREDOARRIETA202082, 8631448, 9233366,Wiegreffe2021TeachMT}). We approach the problem from a slightly different angle: this paper proposes two bottom-up interpretable models for universal approximation, namely triangularly-constructed NN (TNN) and Semi-Quantized Activation NN (SQANN). All codes are available in supp. materials \url{https://github.com/ericotjo001/explainable_ai/tree/master/xaia}.

\textbf{Related works and interpretability issues}. Recent remarkable studies on universal approximators include the Deep Narrow Network by \cite{kidger2020universal}, DeepONet universal approximation for operators by \cite{Lu2021} and the Broad Learning System by \cite{8457525};  \cite{math7100992,park2021minimum,johnson2018deep}. While insightful, they do not directly address the eXplainable Artificial Intelligence (XAI) issue, especially the blackbox property of the DNN. Similarly, a number of classical papers provide theoretical insights for NN as universal approximators, but interpretability, transparency and fairness issues are not their main focus. The universal approximation theorem by \cite{Cybenko1989} asserts that a NN with a single hidden layer can approximate any function to arbitrarily small error under common conditions, proven by asserting the density of that set of NN in the function space using classic mathematical theorems. In particular, its theorem 1 uses an abstract proof by contradiction. From the proof, it is not easy to observe the internal mechanism of a NN in a straight-forward manner; consequently modern works that depend on it (e.g. Deep Narrow Network) might inherit the blackbox property. Bottom-up constructions for function approximation using NN then emerged, though they also lack the interpretability (see appendix \ref{appendix:related works} for more related works). Also consider a demonstration in \cite{nielsen_2015} that could help improve our understanding of universal approximation.

\subsection{Problem and Concept Definitions}
\label{section:definitions}
 \textit{Important Remark}. We submitted the second version of this paper to ICML 2022, and a reviewer kindly commented: ``\textit{since the ideas presented in this paper are entirely novel, the authors should clearly, precisely and unambiguously define everything}", which a meta-reviewer agreed on. Thus, in order to achieve better clarity and precision in presentation, we are compelled to start off by (re)defining multiple terms. Hence, this section will largely be dedicated to disambiguation.

\textbf{Our models have a neural network structure}. We use standard notations for function (and function compositions) to represent our models: \(y=f(x)=a_L\circ\dots \circ a_1(x)\) for some functions \(a_k\). There's nothing special about \(a_k\), except it has an activation function at its output (we use sigmoid and \(d_{dsa}\), to be defined later). The intermediate activation at layer \(k\) is denoted by \(v_k\equiv v_k (x)\equiv f_k(x)=a_k\circ\dots \circ a_1(x)\). 

Our prototype model is \(f=TNN\) which consists of a single layer of \(N\) neurons/nodes (we use neurons and nodes interchangeably). Each neuron is associated to an \(\alpha\) value which is used for the computation of output. For example, when input \(x\) activates the first \(k\) neurons, \(TNN(x)=\Sigma_{i=1}^k \alpha_i := y\). Our main model, \(f=SQANN\), consists of \(L\) layer of neurons. Layer \(k\) has \(n_k\) neurons. Similar to TNN, each neuron has an \(\alpha\) value, and \(SQANN(x):=y\) is computed using possibly more than one values of \(\alpha\) with customizable interpolation functions.

\textbf{Dataset}. We define a \textit{dataset} as a finite ordered set of data samples \(\{(x^{(k)},y^{(k)})\in X\times Y:k=1,\dots,N\}\) where \(X\) denotes the set of input and \(Y\) the set of ground-truth labels or output. We also refer to \((x^{(m)},y^{(m)})\) as sample \(m\). The constructed models are deterministic, dependent on data ordering. A \textit{fitting dataset} \(D\) is a dataset such that each sample \(m\) will be integrated into the model i.e. its values are stored in the \(i\)-th neuron in layer \(l\) for some \(i,l\) e.g. in our main model SQANN, the stored values correspond to node \(\eta_l^{<i>}\) with \(\alpha\) value \(y_l^{<i>}\). An \textit{external dataset} \(D'\) is any other dataset whose sample \((x,y)\notin D\). An element of \(D,D'\) is called a \textit{fitting, external sample} respectively. \textit{Remark}: previously, we use the terms training/validation/test dataset the way they are commonly used in ML community. However, a reviewer finds it contentious: ``\textit{[our setup] is quite an unusual ... in a ML setting}" particularly because we ``\textit{absorb ood test data points}". Hence, we have introduced the following distinct set of terms: fitting/external dataset instead of training/test dataset respectively.

\textbf{Universal Approximation} (UA). Let the parameters of model \(f\) be denoted collectively as \(\theta\). Parameter \(\theta\) comprises any tunable values in \(f\), including \(\alpha,\eta\). In this paper, UA is defined as the perfect approximation of fitting dataset i.e. there exists optimal \(\theta^*\) such that \(\forall (x,y)\in D\) such that \(f(x)=y\). \textit{Remark}. Readers might be familiar with UA of functions with certain common conditions, e.g. compact sets. Our models can be more general, e.g. user can freely choose the interpolation function between two activations based on knowledge of the local manifold. The function can even be pathological; see appendix \ref{appendix:ua} for more descriptions. In our common implementation, no knowledge of any local manifold is included.

\textbf{Absorbing out of distribution data (ood data/samples) into \(D\)}. A sample \((x,y)\) is said to be ood if \(|f(x)-y|>\epsilon\) for a user-determined \(\epsilon\). If \((x,y)\in D'\) is ood, we can extend the fitting dataset \(D\rightarrow D\cup\{(x,y)\}\) such that \(x\) will subsequently be predicted correctly by the new, updated model i.e. our model has \textit{absorbed} the ood sample \(x\). The external dataset is then updated \(D'\rightarrow D'-\{(x,y)\}\). The model can be continually updated until every ood sample in \(D'\) has been absorbed.

\textbf{Catastrophic forgetting} (CF): the tendency for knowledge of previously learned dataset to be abruptly lost as information relevant to a new dataset is incorporated. This definition is a slightly nuanced version of \cite{Kirkpatrick3521}. More precisely, our models' resistance to CF is the following. Suppose \(D\) is used to construct model \(f\), so \(y=f(x)\) for all \((x,y)\in D\). When new samples are used to update \(f\), the new model \(f'\) still predicts all previous fitting samples accurately, i.e. \(f'(x)=y\). See appendix \ref{appdx:cf} (analogy to biological system included).

Our definition of CF might be at odds with readers familiar with training/validation/test divisions of datasets. With the current standard in ML community, validation or test datasets will never be incorporated into training dataset. Let us reiterate that we do incorporate data from \(D'\) into \(D\) \textit{only when they have been ruled as ood}. From our models, ood data can be identified e.g. through weak activations of SQANN neurons or when a prediction yields error \(>\epsilon\).

\textbf{Interpretability}. We consider only fine-grained interpretation, i.e. we look at what each single neuron represents. Our model is interpretable in the following sense: each neuron in the model is identifiable with a particular data sample in the fitting dataset through a pattern of neuron activations. In this paper, we consider mostly patterns with maximal activation, where maximum activation is designed to be \(1\). More precisely, for any \((x,y)\in D\), there exist \(i,j\) with a specific neuron activation property/pattern e.g. for TNN, \(f_j(x)_{i'}=1\) for all \(i'\le i\) and for SQANN \(f_j(x)_i=1\). 

For external samples, interpretability can be achieved, for example, by finding which neuron has the greatest activation. The fitting sample associated to this neuron can be considered as the ``reference data" or ``similar data" that provides the context for interpretability.

\textbf{More on Notations}. Recall that superscript in the form of bracketed index \((k)\) denotes anything related to the \(k\)-th sample e.g. see how a sample \((x^{(k)},y^{(k)})\) in a dataset is defined above. They will be useful to identify the relevant fitting sample in other notations too e.g. the activation pattern for SQANN above can be written with sample order subscript \(\forall k (x^{(k)},y^{(k)})\in D\), \(\exists (i,j)f_j(x^{(k)})_i=1\). 

\textbf{What this paper is about}. Construction of interpretable UA models with three properties: (1) resistance to catastrophic forgetting (2) provably accurate approximation and (3) the ability to identify out-of-distribution (ood) samples in an interpretable way (by recognizing weak activations). 

\textbf{What this paper is NOT about}. Readers/reviewers expecting to read about superhuman feat recently popularized by DNN models will find themselves slightly off the mark. We are not proposing a high-performing latent encoder with remarkable generalization capability. As previously mentioned, generalization is handled \textit{differently}: we absorb ood external data points into fitting dataset. 

\textbf{What is ideally expected?} Recent ML/DNN-centric readers may be expecting good extrapolation or generalization to unseen real-life data. We certainly share the same ideal. However, in this paper, we devise UA models for perfect approximation, not for creating the perfect latent representation. If given a perfect latent representation as the input with correct labels, UA models will extrapolate well. 

Section 2 shows explicit TNN construction, its UA and other properties, including \textit{a pencil-and-paper example} for pedagogical purpose. Likewise, section 3 shows SQANN construction, statements regarding SQANN, \textit{another pencil-and-paper example} and a brief description of experiment on a few small datasets. Most other details are in the appendix. We conclude with the models' shortcomings, future works and ongoing developments.
  
\section{Triangularly-constructed NN (TNN)}
\label{section:TNN}
TNN is our prototype interpretable UA model which uses the following concepts: (1) organized activations of neurons and (2) the retrieval of \(\alpha\) values for output computation. It is a small model given by \(TNN(x)=\alpha^T \sigma(Wx+b)\) where \(x\in [0,1]\), \(\alpha, b\in\mathbb{R}^N\) and \(W\in\mathbb{R}^{N}\); we use sigmoid function \(\sigma\) for simplicity. We start with the approximation of a simple scalar function \(y=f_(x)\in\mathbb{R}\) for \(x\in [0,1]\), thus TNN's interpretability can be illustrated very clearly. This further helps illustrate our main model SQANN in the next section.

\begin{figure}
  \centering
  \includegraphics[width=\columnwidth]{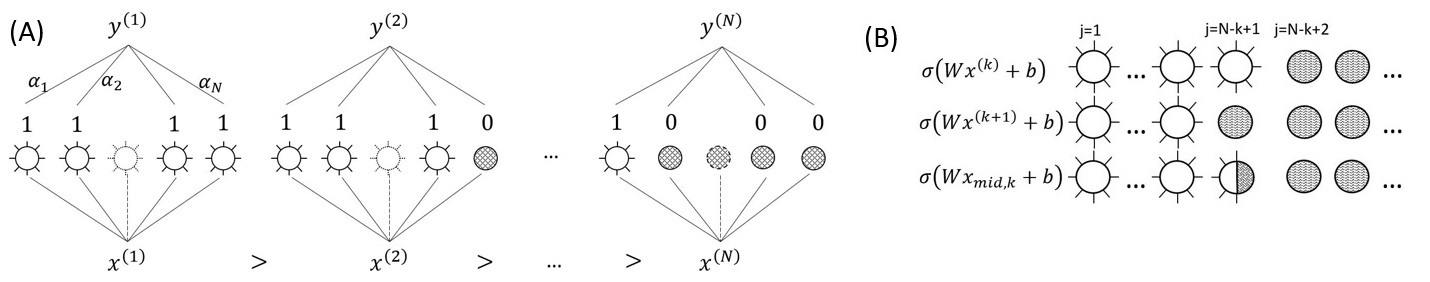}
  \caption{(A) Triangular construction is built by prioritizing interpretability of a neural network. As \(x^{(k)}\) decreases in ``strength", the neurons are ``turned off" correspondingly. (B) Activations of neurons for \(x^{(k)},x^{(k+1)}\) and their mid-point \(x_{mid,k}\). Not only will neuron activation be half at the mid-point, the output \(y_{mid,k}=\frac{1}{2}(y^{(k)}+y^{(k+1)})\) is also half the sum of its neighbours'.}
  \label{triangularcon}
\end{figure}

\textbf{Linear Ordering}. For TNN, further assume that the \textit{fitting dataset} \(D\) is linearly ordered: \(x^{(N)}<x^{(N-1)}<\dots <x^{(1)}\) and \(y^{(k)}=f_0(x^{(k)})\) where \(f_0\) is the true function that TNN will approximate. The interpretability comes from the linear ordering property where higher value of \(x\) (nearer to \(1\)) will activate more neurons while lower values will activate less neurons as shown in fig \ref{triangularcon}(A).  Then \(\alpha\) values will be retrieved in a continuous way through dot product, eventually used to compute the output for prediction. This property can be used for time series e.g. ECG (Electrocardiogram) where signals can be approximated point-wise (although it is still preferable to have a noise model during preprocessing to prevent overfitting the noise). Meaningful interpretation can be given e.g. by mapping PQRST segments from ECG to specific neurons within TNN, giving some neurons specific meaning and thus interpretability. For more remarks and definition of formal \textit{linear ordering} etc, see appendix \ref{appendix:tnn}.

\textbf{Ordered activation}. We would like \(x^{(1)}\) to activate all neurons, while \(x^{(N)}\) activates only 1 neuron. In other words, ideally \(\sigma(Wx^{(1)}+b)=[1,1,\dots,1,1]^T\), followed by \(\sigma(Wx^{(2)}+b)=[1,1,\dots,1,0]^T\) and so on until \(\sigma(Wx^{(N)}+b)=[1,0,\dots,0]^T\); again, refer to fig. \ref{triangularcon}(A). With this concept, we seek to achieve interpretability by the successive activations of neurons depending on the ``intensity" of the input, with \(x^{(1)}\) being the most intense. In general, the above can be written as 
\begin{equation}
\label{maincondition}
\sigma^{(k)}\equiv\sigma(Wx^{(k)}+b)=[\underbrace{1,\dots,1}_{N-(k-1)},\underbrace{0,\dots,0}_{k-1}]^T
\end{equation}
which is approximately achieved for \(k=1,\dots,N\) at large \(a\) (and exactly if \(a\to\infty\)) with 
\begin{equation}
\label{maincondition2}
(Wx^{(k)}+b)_j= \begin{cases} \le -a, j\ge N-k+2 \\ \ge +a \ ,j\le N-k+1 \end{cases}
\end{equation}
For more remarks, see appendix \ref{appendix:tnn} subsection \textit{ordered activation}.

\begin{figure}
  \centering
  \includegraphics[width=\columnwidth]{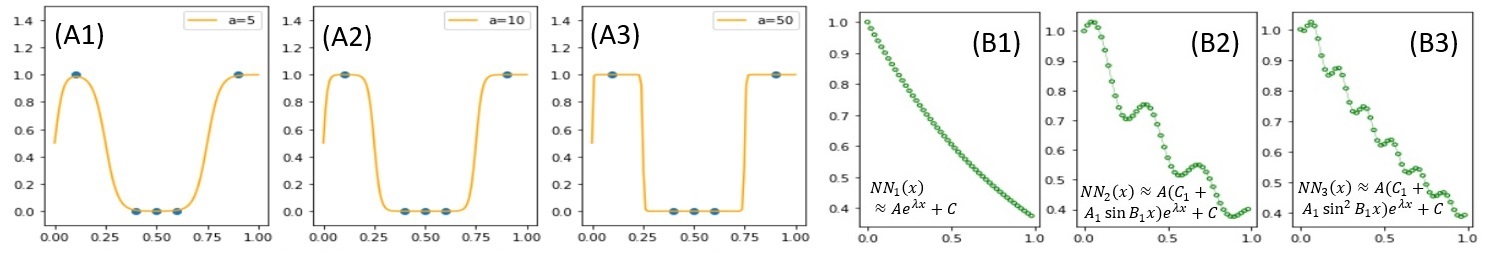}
  \caption{(A1-3) Three triangular constructions (orange plots) using different values of \(a\). Higher \(a\) results in more step-wise plots and more constant values around the data samples (blue points). (B1) Plots of NNs approximated using triangular construction (smooth green plots) over scatter plots of the corresponding true data (green open circles). The parameters are as the following \(A=1, \lambda=-1, C=0\) for all, (B2) \(A_1=0.1, B_1=20, C_1=1\) (B3) \(A_1=0.1, B_1=10, C_1=1\).}
  \label{fig:3exps}
\end{figure}

\textbf{TNN construction: computing weights \(W,b,\alpha\)}. How then do we compute \(W,\ b\) to achieve the \textit{ordered activation}? Consider first \((Wx^{(2)}+b)_N = -a\) and \((Wx^{(1)}+b)_N = a\) and solve them. This yields \(W_N=2a/\Delta^{(1)}\) and \(b_N=a-W_N x^{(1)}\) where \(\Delta^{(1)}=x^{(1)}-x^{(2)}\). Iterating through k, i.e. solving \((Wx^{(k+1)}+b)_{N-k+1} = -a\) and \((Wx^{(k)}+b)_{N-k+1} = a\) we obtain \(W_{N-k+1}=2a/\Delta^{(k)}\) and \(b_{N-k+1}=a-W_{N-k+1} x^{(k)}\) where \(\Delta^{(k)}=x^{(k)}-x^{(k+1)}\). We can rewrite the indices so that \(W_k=2a/\Delta^{(N-k+1)}\) and \(b_k=a-W_k x^{(N-k+1)}\) whenever convenient. For \(\Delta^{(N)}\), we need a dummy \(x^{(N+1)}\) value or we can directly choose its value, e.g. \(\frac{1}{N} \Sigma_{k=1}^{N-1} \Delta^{(k)}\). The effect is illustrated by the value near \(x=0\) in fig. \ref{fig:3exps}(A1-3) and should not pose any problem; the chosen dummy value will only affect the final shape at the left end of the graph. 

We compute \(\alpha\) using the property of equation (\ref{maincondition}). From fig. \ref{triangularcon}(A), this means ideally \(y^{(1)}=\Sigma_{i=1}^{N}\alpha_i \sigma(Wx^{(1)}+b)_i\) for \(a\to\infty\), and similarly \(y^{(2)}=\Sigma_{i=1}^{N-1}\alpha_i \sigma(Wx^{(2)}+b)_i\) and so on until \(y^{(N)}=\alpha_1 \sigma(Wx^{(N)}+b)_1\). Putting them together as \(y=[y^{(1)},\dots,y^{(N)}]^T\), we get \(y=A\alpha\) where \(A\) is an upper-left triangular matrix and the inverse \(A^{-1}\) exists. Thus, \(\alpha=A^{-1}y\), a matrix such that \(A_{ij}^{-1}=1\) along the diagonal, \(A_{i,i+1}^{-1}=-1\) and zeroes otherwise, which facilitates a convenient computation. The triangular construction is complete:
\begin{equation}
\label{eq:form}
TNN(x)=\alpha^T\sigma(Wx+b)
\end{equation} 

While \cite{nielsen_2015} provides only visual demonstration, our result below shows a rigorous proof of UA at work. Python code is available; the code (1) shows zero error on \(D\) and (2) plots example like fig. \ref{fig:3exps}. Note that user-specified tolerance \(\epsilon\) is specified as the upper bound of error in the following UA.
\begin{theorem}
TNN is a universal approximation of \(D\). Proof: see appendix \ref{theorem1proof}. 
\label{eq:tnn_acc}
\end{theorem}

External dataset \(D'\) that resembles fitting dataset \(D\) is expected to be approximated well (small error). Otherwise, there exist ood samples \(A\subseteq D'\) with large error. How large the error is depends on user's tolerance; for convenience we use the \(\epsilon\) from the UA theorem. With the following proposition, ood samples are absorbed into fitting dataset \(D\) so that the model learns to distinguish new data correctly without forgetting old samples. In other words, the model does not suffer from CF after re-training because each sample in \(D\) will still be represented by exactly one neuron after absorption \(D\rightarrow D\cup A\), obviously because \((x,y)\in D\) still holds. Hence, by theorem \ref{eq:tnn_acc}, \(x\) is still approximated accurately. See appendix \ref{appdx:cf} for remarks on \textit{advantage over existing methods}.

\begin{proposition}
\textbf{TNN does not suffer from CF}. Let fitting and external datasets be \(D,D'\). Let the user specified tolerance be \(\epsilon\). There exists \(A\subseteq D'\) such that if TNN is constructed with \(D\cup A\), then for any \((x,y)\in D\cup D'\), sample-wise error \(e=|y-TNN(x)|<\epsilon\). Proof: see appendix \ref{proof:errorbound}.
\label{prop:errorbound} 
\end{proposition}

\textbf{TNN pencil-and-paper example}. Use TNN to fit the dataset \((x,y)\in\{(1,1),(0.5,2),(0,3)\}\). Then \(f(x)\approx TNN(x)=3\sigma(20x+5)-\sigma(20x-5)-\sigma(20x-15)\). See appendix \ref{TNNexample}.

\textit{Remarks}. TNN has a mid-point property: the mid-point \(x_{mid,k}=\frac{1}{2}(x^{(k)}+x^{(k+1)})\) takes the value of \(\frac{1}{2}(y^{(k)}+y^{(k+1)})\). Smoothness property, special case, scalability/complexity and generalizability to higher dimensions can also be found in appendix \ref{tnnremarks}. We proceed with SQANN, a universal approximation inspired by TNN that allows multi-dimensional input, multi-layer stacking of neurons based on relative strength of neuron activations.

\begin{figure}
  \centering
  \includegraphics[width=\columnwidth]{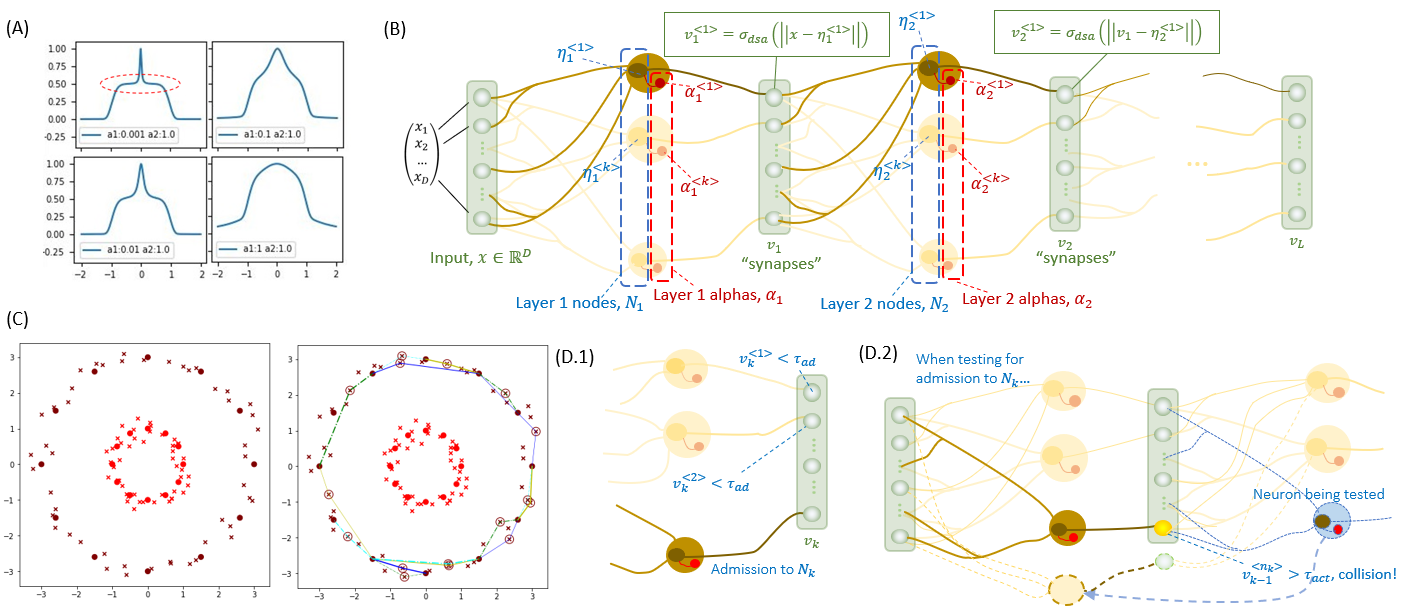}
  \caption{(A) Double selective activation with different parameters. Red area shows a distinct ``moderate" activation. (B) SQANN schematic. Each layer \((N_k,\alpha_k)\) is stylized as a collection of neurons. A neuron stores the main ``fingerprint"  in nucleus \(\eta_l^{<k>}\) (dark brown) and its corresponding ``output" in the alpha nucleus \(\alpha_l^{<k>}\) (dark red). When strong activation is detected, the signal will be redirected to the alpha nucleus \(\alpha_l^{<k>}\). (C) SQANN used for a simple classification. Left: the large filled dots are fitting samples, x marks are external samples. Bright red points represent samples wiht \(y=1.0\), dark red \(y=0.5\). Right: same as left except external samples that are interpolated (i.e. no strong activation) are annotated with red open circles; colored lines indicate which two fitting samples are used for the interpolation. Lines are marked with different colors and styles for clarity. (D) Construction of SQANN when (D.1) admission occurs: a new neuron is introduced, creating more connection analogous to mammalian brains. (D.2) collision occurs.}
  \label{fig:sqann}
\end{figure}

\section{Semi-Quantized Activation Neural Network (SQANN)}
\label{section:SQANN}
SQANN architecture is a multi-dimensional UA which (1) retains TNN's idea of using an organized sequence of activations to retrieve \(\alpha\), (2) remotely resembles a Radial Basis Function, but (3) has DNN properties, e.g. deep learning (multiple constructed layers) and neuron activations. Due to its non-linear multi-layer structure, SQANN is a generalizing model (by contrast, nearest neighbours methods have been considered non-generalizing; see the remark in \cite{scikit-learn}).  

Difference from DNN: a neuron in SQANN corresponds exactly to a sample \(m\): the model stores the ``fingerprint" of sample \(m\) as a neuron's nucleus \((\eta,\alpha)\) as shown in fig. \ref{fig:sqann}(B). These neurons can be activated with three kinds of responses: (1) distinct peaks and (2) half-activations and (3) weak/zero activations, made possible by double selective activation \(\sigma_{dsa}\). They are illustrated in fig. \ref{fig:sqann}(A) and later results. 

The distinct types of activation are key to SQANN's main result and give the model an interpretability at least in the following sense: samples are highly/moderately/not recognizable if their activation patterns strongly/moderately/weakly resemble the activation patterns of an existing fitting sample \(x\). The different strengths of activations arevmade possible by the distinct regions of \(\sigma_{dsa}\), e.g. moderate activation is highlighted in fig. \ref{fig:sqann}(A) with red mark. The design is ``semi" quantized since \(\sigma_{dsa}\) has approximately ``distinct" levels yet remains a continuous function. 

\textbf{Notations}. Intermediate activation \(v\) (stylized as the ``synapse") is indexed as follows: subscript indicates layer, square bracket with subscript denotes vector component so that \([v_2]_4\) is the 4-th vector component of the activation of layer 2. \textit{Layer k} consists of \((N_k,\alpha_k)\), where \(N_k=(\eta_k^{<1>},\eta_k^{<2>},\dots,\eta_k^{<n_k>})\) stores fingerprints/patterns, \(\alpha_k=(y_k^{<1>},y_k^{<2>},\dots,y_k^{<n_k>})\) stores output values. \textit{Angle brackets} denote indices after relabeling. Hence, if sample \(k=50\) becomes the first node in layer 2, then we write \(\eta_2^{<1>}=\eta_2^{(50)}\). \textit{Concatenation}. To denote the addition of the new \(k\)-th node to the layer \(l\), use \(\eta_l^{<k>}\leftarrow v\), where \(v\) can be for example \(v_2^{(m)}\) the activation of the \(m\)-th data at layer 2. Alternatively, \(N_l\rightarrow concat(N_l,\eta_l^{<k>})\). We can speak about layer \(k\) using \(N_k\) without mentioning \(\alpha_k\) when it is not involved. However, once concatenation of \(N_l\) is decided, concatenation of \(\alpha_l\) always follows i.e. \(y_l^{<k>}\leftarrow y^{(m)}\). 

\textbf{Selective clustering} of \(p_k=(x_k,y_k)\) for \(k=1,2\) is loosely defined for \(x_1,x_2\) that are close to each other such that: if \(y_1,y_2\) are similar (case 1), then \(p_1,p_2\) are clustered together. Otherwise (case 2) two distinct clusters with different labels/outputs/alphas \(y_1,y_2\) are created. See appendix \ref{appendix:selectiveclustering} for formal definition, its effects on interpolation and more remarks.

\textbf{Double selective activation}. Given selective activation \(\pi (x,a)=\frac{a}{a+x^2}\) and Super Gaussian \(s_g(x, a)=exp(-(x/a)^{8})\), \(r=0.5\), then the \textit{double selective activation} is (fig. \ref{fig:sqann}(A)): 
\begin{equation}
\sigma_{dsa}(x,a_1,a_2,r)=(1-r) \times\pi(x,a_1)+ r\times s_g(x,a_2) \\
\label{eqn:dsa}
\end{equation}
\textbf{Nodes activation}. Denote the ``synapse" or the activation value of node \(j\) at layer \(k\) by input \(v\) as:
\begin{equation}
[v_k]_j=\sigma_{dsa}\big( ||v-\eta_k^{<j>}||\big)
\label{eqn:act}
\end{equation}
where \(\eta_k^{<j>}\in N_k\) for \(j=1,\dots,n_k\) and \(n_k\) is the number of neurons/nodes in the layer. See fig. \ref{fig:sqann}(B). In SQANN, activations will be forwarded layer by layer, i.e. \([v_1]_j=\sigma_{dsa}\big( ||x-\eta_1^{<j>}||\big)\) where \(\eta_1^{<j>}\in N_1\) and \([v_{k+1}]_j=\sigma_{dsa}\big( ||v_{k}-\eta_{k+1}^{<j>}||\big)\) where \(v_{k+1}^{<j>}\in N_{k+1}\). 

\subsection{SQANN Construction}
\textbf{Outline of SQANN construction with interpretations}. SQANN is constructed without optimization like gradient descent. Each indexed fitting data sample is converted into a ``fingerprint" or pattern of neuron activations, which undergoes one of the following: 
\begin{enumerate}[leftmargin=*,topsep=0pt]
\item \textit{Admission to layer \(k\)}. Sample's new/distinct fingerprint is added into layer \(k\) if the sample weakly activates existing nodes in the layer (\(\forall j,[v_k]_j<\tau_{ad}\)) and no collision occurs; see fig. \ref{fig:sqann}(D.1).
\item \textit{Collision}. A sample activates one or more existing neurons strongly i.e. \(\exists  j,k, [v_k]_j>\tau_{act}\). Denote the earliest layer where collision occurs as \(l_c\). Then such collided sample will be integrated into \(l_c\). In other words, two samples \((x_k,y_k)\), \(k=1,2\) with similar \(x_1,x_2\) will be \textit{selectively clustered} through the reconstruction of layer \(l_c\). See fig. \ref{fig:sqann}(D.2). The concept is also illustrated in the sketch of proof for proposition \ref{prop:complete}, appendix \ref{appendix:sqanncompleteproof}.
\item \textit{Filtering into deeper layer} occurs when neither of the above occurs (no strong activation, some moderate activations). Such sample has features loosely similar to previously seen samples, but we need to filter them further to distinguish their finer features. 
\end{enumerate}

\textbf{Layer 1 construction}. To initialize, let \(N_1=(\eta_1^{<1>})\) and \(\alpha_1=(y_1^{<1>})\) where \(\eta_1^{<1>}\leftarrow x^{(1)}\) and \(y_1^{<1>}\leftarrow y^{(1)}\). Let \(\tau_{ad},\tau_{act}\) be the \textit{admission threshold} and \textit{activation threshold} (typically set to \(0.1, 0.9\)) respectively. Then, we extend the layer to tuples \(N_1=(\eta_1^{<1>},\dots,\eta_1^{<n_1>})\) and \(\alpha_1=(y_1^{<1>},\dots,y_1^{<n_1>})\) through sample-collection function (pseudo code \ref{code:sqann}): take a new sample \((x^{(k)},y^{(k)}),k\ge 2\). Then we \textit{check \(N_1\) activation} to see which of the three conditions hold (see outline of SQANN construction above) as the following. Let \(v_1^{(k)}\) be the activation of current layer by this new sample, i.e. \([v_1^{(k)}]_j=\sigma_{dsa}\big( ||x^{(k)}-\eta_1^{<j>}||\big)\) for all \(\eta_1^{<j>}\in N_1\). Then, either: 
\begin{enumerate}[leftmargin=*,topsep=0pt]
\item Admission to layer \(1\). If for all \(j\) such that \([v_1^{(k)}]_j<\tau_{ad}\), then \(N_1\rightarrow concat(N_1,x^{(k)})\) and \(\alpha_1\rightarrow concat(\alpha_1,y^{(k)})\) i.e. new sample is \textit{admitted to \(N_1\)} as a new distinct node/neuron. 
\item \textit{collision} occurs, when there exists \(j\) such that \(\tau_{act}<[v_1^{(k)}]_j<1\). Unlike later layers \(l>1\), sample will simply be admitted into layer 1 as a node; selective clustering occurs.  
\item or new sample is \textit{filtered to deeper layers}, when neither occurs. 
\end{enumerate}
Finally, we complete the iteration over all fitting data, \(N_1=(\eta_1^{<k>}:k=1,\dots,n_1)\) and \(\alpha_1=(y^{<k>}:k=1,\dots,n_1)\). Note: new sample causing \([v_1^{(k)}]_j=1\) collision is unresolvable; see appendix on ill-defined datasets.

\begin{lemma}
Layer 1 of SQANN achieves UA on fitting data subset \(N_1\times\alpha_1\).
\label{lemma:first}
\end{lemma}
Proof: Let \(N_1=(x^{<1>}=x^{(1)},x^{<2>},\dots,x^{<n_1>})\); note that \(x^{<k>}\) is not necessarily \(x^{(k)}\) except for \(k=1\) for initialization. To prove the lemma, take a sample \((x,y)\in N_1\times\alpha_1\). Then we must have \(x=x^{<j>}, y=\alpha_1^{<j>}\) for some \(j=1,\dots,n_1\). Since \([v_1]_{j'}=\sigma_{dsa}(||x-\eta_1^{<j'>}||)\) and \(\eta_1^{<j'>}=x^{<j'>}\) for all \(j'=1,\dots,n_1\), we get exactly \([v_1]_j=1\). Furthermore, for other \(i\neq j\), we have \([v_1]_i<1\) due to the admission conditions (1) and (2) used during \textit{check \(N_1\) activation} process. Finally, computing \(y=\alpha_1^{<j>}\) where \(j=argmax_{j'}[v_1]_{j'}\), we retrieve the exact value. \(\square\)

At this point, it may have become clearer to readers how SQANN is constructed. In short, for each layer, \textit{representative activations} become the neurons of the layer. In layer 1, representatives activations are the samples themselves. In deeper layers, they are activations ``fingerprints" propagated to the layer. 

\lstset{language=python}  
\captionof{lstlisting}{Pseudo code for the construction of SQANN. The function \textit{activate} corresponds to equation \ref{eqn:act}. See appendix \ref{appendix:pseudocode} for mapping to python code.
}
\begin{minipage}{.5\textwidth}
\begin{lstlisting}[columns=fullflexible,]

def fit_data(X,Y):
  # Main SQANN loop
  l_now=1 # layer now
  while True:
    ssig, collision = \
      sample_collection(X, Y, l_now)
    if ssig is 'no more data':
      break
    elif ssig is 'collision':
      l_c = collision['collided_layer']
      for l_j from l_c+1 to l_now+1:
        return_index(l_j)
      kp = collision['perpetrator_index']
      push_node(kp,X[kp,:],Y[kp],l_c)      
      l_now = l_c 
  l_now+=1 
\end{lstlisting}
\end{minipage}\hfill
\begin{minipage}{.5\textwidth}
\begin{lstlisting}[columns=fullflexible]
def sample_collection(X,Y,layer):
  i=unused_indices[0]
  x=X[i,:]
  x, collision = forward_cons(x,layer-1)
  ssig, collision = check_signal(collision)
  nodes, node_values = new_nodes(x,Y[i])       
  remove_index(i,layer)
  for i in unused_indices:
    x=X[i,:]
    x, collision = forward_cons(x,layer-1)
    ssig, collision = check_signal(collision)
    act=activate(x,nodes) 
    if all(act<admission_threshold)     
      update_nodes(x,Y[i],nodes,node_values)     
      remove_index(i,layer)
  return ssig, collision
\end{lstlisting}
\end{minipage}
\label{code:sqann} 
\parskip0.5em

\textbf{Layer k construction}. Layer \(k\) construction is similar to layer 1 construction, except collision could occur at any layer \(l_c\le k\) (next paragraph). Assume every layer \(l\in\Lambda=\{1,\dots,k-1\}\) have been constructed using \(X_{i\in\Lambda}\subseteq X\). Assume there are still unused data samples i.e. \(U=X\setminus\{\bigcup_{i=1}^{k-1} X_{i}\}\) is non-empty, obtained from samples that have been \textit{filtered to deeper layers}. Let \(U=\{u^{<1>},u^{<2>},\dots\}\) after re-labelling the indices, with corresponding output values \(\{y^{<1>},\dots\}\). Initialize by first checking \(v_k^{<1>}\), the activation of \(u^{<1>}\) at layer \(k\) for collision. Assume no collision occurs here, set \(N_k=(\eta_k^{<1>}),\alpha_k=(y_k^{<1>})\) i.e. \(\eta_k^{<1>}\leftarrow v_k^{<1>}\). Similar to layer 1 construction, perform \textit{check \(N_k\) activation} on \((u^{<i>}, y^{<i>})\) for \(i>1\) by computing activation \(v_k^{<i>}\) and checking it against the existing nodes. One of the three cases occur (1) admission, when \([v_k^{<i>}]_j<\tau_{ad}\) for all \(j=1,\dots,n_k\) and no collision has occurred (2) collision, when there exists index \(j\) at a \textit{collided layer} \(l_c\le k\) such that \([v_{l_c}^{<i>}]_j>\tau_{act}\) or (3) otherwise. 

As before, if (1) occurs, the activation \(v_k^{<i>}\) is added as a new neuron to the layer, \(N_k\rightarrow concat(N_k,v_k^{<i>})\). If (3) occurs, filter the data sample for deeper layer. Assuming no collision, the process is repeated for the next unused data sample \(u^{<i>}\) until all remaining data samples are checked. Once done, repeat the process for layer \(k+1\) construction.

Suppose collision happens anytime during the above process, i.e. \(u^{<m>}\) triggers activation \(>\tau_{act}\) at some layer \(l_c\). We use the collision resolution mechanism as the following. Destroy all layers \(l>l_c\) and put the collided sample into \(l_c\), i.e. \(N_{l_c}\rightarrow concat(N_{l_c},v_k^{<m>})\) and \(\alpha_{l_c}\rightarrow concat(\alpha_{l_c},y_k^{<m>})\) (push-node in the pseudo code). No layer will be destroyed if \(l_c=k\), the current layer. The data samples used in each destroyed layer are returned to the list of unused samples in the same order they have been used during the construction (return-index() in the pseudo code); we refer to this as \textit{order integrity}. Once the colliding sample is added to \(l_c\), the construction process proceeds: nodes that have been destroyed will be reconstructed, and new nodes beyond \(u^{<m>}\) will undergo \textit{check \(N_k\) activation} as well. The profile of the resulting layers up to \(u^{<m>}\) is expected to be similar to before since the nodes are added one after another in the same order (not exactly the same because now \(u^{<m>}\) node is present). 

The strong activation that caused collision in layer \(l_c\) is now effectively overshadowed by maximum activation \([v_{l_c}]_i=1\) for some \(i\) (\textit{selective clustering} in action) since the colliding sample has now been included as a neuron that represents itself and its locality. Finally, the correct \(\alpha\) value will then be fetched by the following SQANN propagation mechanism.

\textbf{Computing output via SQANN propagation (prediction)}. Let an input be \(x\). The output \(y=SQANN(x)\) is computed by propagating and processing signals through the layers; see fig. \ref{fig:sqann}(B).  Then \([v_1]_j=\sigma_{dsa}\big(||x-\eta_1^{<j>} ||\big)\). If there exists \(j\) such that \(v_1^{<j>}>\tau_{act}\), then set \(y=\alpha_1^{<j>}\) where \(j=argmax_{j'}[v_1]_{j'}\). Otherwise, for subsequent layer \(k\), recursively compute \([v_k]_j=\sigma_{dsa}\big(||v_{k-1}-\eta_{k}^{<j>} ||\big)\) for all \(j=1,\dots,n_k\). If there exist \(j,k\) such that \([v_k]_j>\tau_{act}\), then \(y=\alpha_k^{<j>}\) where \(j=argmax_{j'}[v_k]_{j'}\). If such layer \(k\) is not found, we have to perform interpolation. 

Interpolations can be done in many different ways, and this paper implements a simple interpolation using values from the two most strongly activated neurons. Suppose \(V_1=[v_m]_i\) and \(V_2=[v_n]_j\) are the two most activated neurons, then the interpolated value can be, for example, \(y=\frac{V_1[\alpha_m]_i +V_2[\alpha_n]_j}{V_1 +V_2}\). The form of interpolation can be adjusted according to the knowledge we have on the dataset, e.g. we can use TNN with high \(a\) if we know the dataset is locally constant. See appendix \ref{appendix:sqannprop} for illustration and remarks.

Each fitting sample \(x\) admitted to layer \(N_k\) leaves a fingerprint \(\eta\) such that SQANN propagation of \(x\) gives rise to activations \(\{v_l|l=1,\dots,k\}\) which is unique by design, especially because of \([v_k]_j=1\) where \(j\) is an index within layer \(k\) it is admitted into. Furthermore, locality is preserved to the extent that if \(x'\approx x\), then activation \([v'_k]_j\approx 1\) i.e. \(x'\) is found near the vicinity of \(x\)'s cluster in the space of activations. Thus, by SQANN propagation, due to argmax, the model is likely to retrieve \(y\), the ground-truth value corresponding to \(x\). For now, we only use argmax so \(y'=y\). More subtle adjustment can be done to obtain \(y'\approx y\) but \(y'\neq y\) for \(x'\) by incorporating information about the manifold at that locality, if available. See appendix \ref{appendix:scalability} on \textit{good practice for scalability}.

\textbf{Completing construction}. During collision, layers are torn down and reconstructed. Readers might thus wonder if the construction will complete at all. Suppose during layer \(k\) construction, collision occurs at layer \(c\) for \(c<k\). Upon reconstruction back to layer \(k\), layer \(c\) may be torn down again in the next collisions. Is it possible that collision occurs infinitely cyclically? The following proposition addresses the concern through \textit{order integrity} previously mentioned. 
\begin{proposition}
If \(D\) is \(p^{c}_u\)-sparse and \(n\)-redundant, then SQANN construction completes with at least \((1-p^{c}_u)^{n/|D|}\) probability. If there is no complication, SQANN construction completes. Proof: see appendix \ref{appendix:sqanncompleteproof}; definition of sparsity and redundancy are also in the appendix.
\label{prop:complete}
\end{proposition}

\textbf{UA on \(D\)} is relatively simple to prove in the following theorem: roughly for each \((x,y)\in D\), there exist \(l,k\) such that \(x\) maximally activates the node \(\eta_l^{<k>}\), thus the correct \(y\) is guaranteed to be fetched from \(\alpha_l\). CF resistance is proven similarly: when new samples are used for training, previous samples are not forgotten since SQANN stores the particular fingerprint \(\eta_l^{<k>}\) of each sample. Note: Our code provides experimental demonstrations showing zero errors on all fitting samples; example results in fig. \ref{fig:sqann}(C).

\begin{theorem}
\label{theorem:sqann}
 Assume SQANN construction is completed. SQANN is a UA on \(D\). Furthermore, it is resistant to CF. Proof: see appendix \ref{proof:theoremsqann}. 
\end{theorem}

\textbf{SQANN pencil-and-paper example}. With \(a_1,a_2=0.001,0.5\) and \(\tau_{ad},\tau_{act}=0.1,0.9\), create SQANN universal approximator for indexed data \(X=[x^{(1)},x^{(2)},x^{(3)},x^{(4)}]=\big[\begin{smallmatrix}
1 & 1.2 & -1 & -1.2\\1.2 & 0.8 & -1 & -1.2\end{smallmatrix}\big]\) and \(Y=[y_1,y_2,y_3,y_4]=[1,1,0,0]\). See appendix \ref{appendix:sqannexample} for more questions and solutions.

\begin{figure}
  \centering
  \includegraphics[width=\columnwidth]{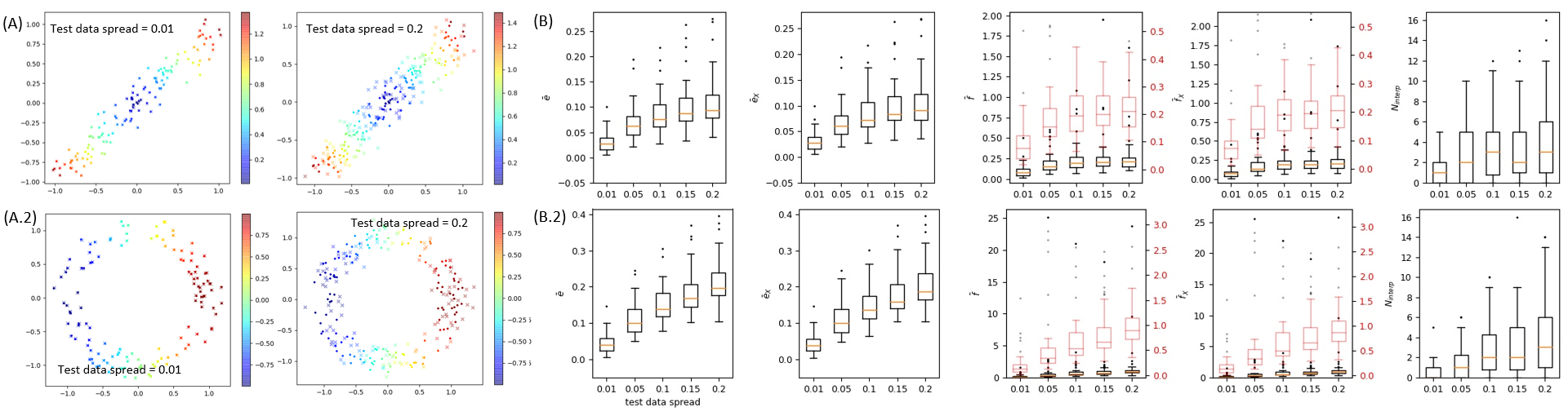}
  \caption{(A) Fitting/external (circles/x marks) datasets for demonstration. Smaller/larger \textit{test data spread} means external samples are closer/further to/from fitting samples. (B) Boxplots for data whose distributions are similar to (A). Column 1(3): (fractional) errors on external samples. Column 2(4), (fractional) errors on external samples excluding interpolated samples.  Column 5: no. of data samples whose predicted values are interpolated. (B.2) Similar to top, but for (A.2).}
  \label{fig:sqann_expt}
\end{figure}

\subsection{Experiment to test Generalizability of SQANN}
\textbf{External datasets with increasing spread from fitting dataset distribution}. The accuracy of SQANN on high-dimensional dataset outside the fitting dataset is harder to formalize in theorems. Furthermore, real life data is often noisy and possibly not regularly structured. Hence, for SQANN, we avoid making any statements or any assumptions related to external dataset distributions. Some results on simple datasets are shown in fig. \ref{fig:sqann_expt}. We test SQANN on small diabetes and Boston Housing datasets as well; most details are in the appendix \ref{appendix:exptdata} due to the lack of space.

\section{Conclusion}
Finally, we have introduced two UA models with the aforementioned properties. \textbf{TNN} only applies to a dataset with linearly ordered input. However, it can be useful on different types of time series data. \textbf{SQANN} limitation currently include 1) simple sequential drawing of samples that may result in the imbalance of layer size 2) layer destruction during collision resolution might be an inefficient mechanism. In our ongoing work, layer size is fixed, though we weakens the UA property.

For future works, the ideal mentioned in the introduction can be considered. (1) We may test the model on input that is already processed by an external encoder with good latent encoder. In our ongoing work, a model modified from SQANN applies this idea to fetch ``influential data" (2) We may carefully select and order representative data points for better performance. Layers can be constructed with meaningful and purposeful arrangement e.g. deeper layers can be purposefully reserved for rare cases. In conclusion, we have proposed TNN and SQANN, two interpretable NNs for universal approximation designed with three aforementioned properties.

\begin{ack}
This research was supported by Alibaba Group Holding Limited, DAMO Academy, Health-AI division under Alibaba-NTU Talent Program. The program is the collaboration between Alibaba and Nanyang Technological University, Singapore.

%
%
\end{ack}

\bibliography{xaia_neurips_2022}
 \bibliographystyle{unsrtnat}

\medskip

{
\small

\appendix

\section{Appendix}

\parskip0.5em
\subsection{Algorithm Code and Supplementary Materials}
All the codes are available in the supplementary material, and the link to the repository will be made public if the paper is accepted. README.md lists the necessary commands to obtain the results found in this paper. Results in Jupyter notebooks format are also included. Hyperlink back to main text introduction here, section \ref{section:definitions}.

\subsection{Related works}
\label{appendix:related works}
Older works on the construction of universal approximators do not focus on interpretability as well, thus readers have to observe for themselves the shape of the resulting networks and infer meanings out of the components used in the construction. For example, there are works related to spline functions like \cite{MHASKAR1992350, Mhaskar1993, 471870, 10.2307/2153285}; another example by \cite{SartoriAntsaklis} where two-layer NN can be formed by arbitrarily choosing the weights of first layer and computing the weights of second layer explicitly; and section 5 of \cite{Pinkus1999ApproximationTO} shows the equations used to obtain weights and proves the existence of such solutions. In some of them and other related works such as \cite{6796304,Chui1996}, the focus lies in error quantification.

\subsection{Problem and Concept Definitions}
\subsubsection{Universal approximation} 
\label{appendix:ua}
In this paper, UA by TNN and SQANN is not restricted to specific conditions. Traditional approximation may require strong conditions e.g. \cite{Cybenko1989} shows result on \(C(I_n)\), space of continuous function. Continuous function that approximates two points necessitate continuous interpolation for any point along the path between these two points. However, hypothetically, pathological set of data points may exist where the value is not even continuous along the path. Our SQANN, for example, can handle this since we allow user to choose the interpolation function in the case the model opts for interpolations for output computation (e.g. during weak activations).

\subsubsection{Catastrophic Forgetting} 
\label{appdx:cf}
\textbf{CF and online learning}. Continual or online learning have posed some known challenges. In \cite{Kirkpatrick3521},  CF is defined as ``the tendency for knowledge of previously learnt task(s) (e.g. task A) to be abruptly lost as information relevant to the current task (e.g. task B) is incorporated". In our context, we can distinguish task A from task B as old and new dataset. In traditional models, learning from the new dataset which possibly has a different distribution from the old dataset might degrade the performance of the model.

\textbf{Analogy to biological system and resistance to catastrophic forgetting}. Unlike artificial NN, mammalian brain retains old information when it learns new information \textit{by protecting previously acquired knowledge in neocortical circuits}; see \cite{Kirkpatrick3521} and the references thereof. As rats learn new skills, the volumes of dendritic spines in their brains increase (\cite{Yang2009}) while existing dendrites persist, thus they retain old memories.  Both TNN and SQANN have similar property. Particularly, SQANN increases the size of a layer as it progressively acquires new samples during construction (learning). This inevitably increases the number of weights that connect the layers; see fig. \ref{fig:sqann}(D.1, D.2).

Furthermore, they exhibit \textit{resistance to catastrophic forgetting}. In TNN, this is simply because the old sample \(x_{old}\) still yields activation with signature \([1,1,...,1,0,0,...,0]\), in which the last activated neuron (taking the value 1) is still identified with the old sample. For SQANN, each old sample is not forgotten since its exact ``fingerprint" is already registered (i.e. input \(x_{old}\) has been converted) to a neuron's nucleus \(\eta_l^{<k>}\) for some \(l,k\), as shown in fig. \ref{fig:sqann}(B). The activation pattern of a sample \(x_{old}\) includes the value \(1\) in a specific node of a ``synapse", \(v_l^{<k>}\) and the particular combinations of values in other synapses.

\textbf{Advantage over existing methods}. Existing ML models do not typically admit ood samples easily; i.e. even if we re-train the models on the new ood samples, they may end up with either poor results on the same ood data or degrade the prediction performance on previous training samples. As an illustration, a linear regression that includes ood samples may shift the gradient a little, but ood sample can still be far from the regression line. To make the matter worse, when there are many ood enough to change the distribution of the data sample significantly, the model may \textit{forget} the previous distribution: we consider this an instance of CF too.

\subsection{Appendix for TNN}
\label{appendix:tnn}
Link back to main text section \ref{section:TNN}.

The traditional definition of NN with a single hidden layer is given by \(\Sigma_{i} \alpha_{i} \sigma (y_{i}^{T}x+b_i)\) following equation (1) from \cite{Cybenko1989}, with \(x, y_{i}\in \mathbb{R}^n, b_i\in\mathbb{R}\) where \(\sigma\) is any sigmoidal function with the property specified in the paper. A familiar example of a sigmoidal function is the sigmoid function \(1/(1+e^{-x})\). Compared to the traditional version, TNN has a slight generalization on the weights \(W\). The \(W\) used here is also a form that has been used in modern DNN implementation. 

\textbf{Linear ordering}. Linear order is any binary relation with (1) reflexivity (2) transitivity (3) anti-symmetricity (4) \(x\le y\) or \(y\le x\). As is customary (\cite{471870}), the domain of the function is limited to \([0,1]^n\) where \(n\) is the number of dimensions of the input, justifiable for practical dataset with finite domain easily scaled to \([0,1]\). Some readers might raise the purpose of fitting a model to a linearly ordered data. Furthermore, if linear ordering is done by human, they might suggest that interpretability is not improved since the human user already knows about the ordering. To answer that, we again refer to the ECG example we gave earlier: time series is a naturally linearly ordered data and meanings assigned to particular points in time are useful human-constructed concept.

\textbf{Ordered activation}. Recall that we would like \(x^{(1)}\) to activate all neurons, while \(x^{(N)}\) activates only 1 neuron, so that eventually we achieve something like fig. \ref{triangularcon}(A). To approximately fulfil the \textit{ordered activation} conditions stated in the main text, we first need \((Wx^{(1)}+b)_j\ge a\) for all \(j\), where sub-script \(j\) denotes the \(j\)-th component in the vector, to distinguish from superscript \((k)\) which denotes the \(k\)-th data sample according to the linear ordering. The next iteration will be \((Wx^{(2)}+b)_j \le -a\) for \(j=N\) and \((Wx^{(2)}+b)_j \ge a\) for \(j=1,\dots,N-1\), and similarly for other \(x^{(k)}\) for \(k=3,\dots,N\).  With this, we attain eq. (\ref{maincondition}) and (\ref{maincondition2}). 

Since we use sigmoid function, these conditions are ideal and not strictly attainable, because sigmoid function asymptotically achieves 0 and 1 at infinities. Nevertheless, we show later that we can achieve arbitrarily small error \(\epsilon\) by adjusting \textit{activation threshold} \(a\). We define the activation threshold to be the value \(a\) such that \(\sigma(a)=1-\delta\), where \(\delta\) is a small number. For this paper, fixing \(a=5\) is sufficient. The shape of sigmoid function is convenient enough to be symmetrical in the sense that \(\sigma(-a)=\delta\) since \(1/(1+e^{-a})=1-\delta\) can be rearranged to \(1/(1+e^{-(-a)})=\delta\), useful for the proof later. 

\subsubsection{TNN theorem proof}
\label{theorem1proof}
\textbf{Theorem \ref{eq:tnn_acc}}. TNN is a universal approximation of \(D\).

Proof: We need equations (\ref{maincondition}) and (\ref{maincondition2}). In practical situation, where \(a\) is finite, we therefore have \(\sigma^{(k)}_j \ge 1-\delta\) for \(j\le N-(k-1)\) and \(\sigma^{(k)}_j \le \delta\) for \(j>N-(k-1)\) and \(\delta>0\). Define sample error as \(e^{(k)}=|y^{(k)}-[A^{-1}y]^T \sigma^{(k)})|\) and average error per sample will be \(e=\frac{1}{N}\Sigma_{i=1}^{N} e^{(k)}\). Then \(e^{(k)}\le \delta (N+1)U\) where \(U=\max_{k}{|y^{(k)}|}\) is the upper bound for the absolute value of the function over all samples (see proof below). Hence, setting \(\delta=\frac{\epsilon}{U(N+1)}\) guarantees that \(e\le \epsilon\). Since \(\delta\) can be monotonously decreased by increasing \(a\), we have shown that arbitrarily small error \(\epsilon\) can be achieved in this approximation; see fig. \ref{fig:3exps}(B1-3) for plotted examples. Note that \(e^{(k)}=0\) iff \(\delta=0\) iff \(a=\infty\). 

Show that \(e^{(k)}\le \delta (N+1)U\) where \(U = \max_{k}{|y^{(k)}|}\). We abbreviate \(\sigma^{(k)}\) as \(\sigma\), fixing k.

\begin{flalign*}
\begin{rcases}
e^{(k)} = |y^{(k)}-(A^{-1}y)^T \sigma^{(k)}| = & \Big|y^{(k)}-\Big(y^{(N)}{\sigma_1 } + (y^{(N-1)}-y^{(N)}){\sigma_2} + \dots \\
& + (y^{(k+1)}-y^{(k+2)}){\sigma_{N-k}} \\
& + (y^{(k)}-y^{(k+1)}){\sigma_{N-k+1}} \\
\end{rcases}\text{$\sigma_j \ge 1-\delta$}\\
\begin{rcases}
&+ (y^{(k-1)}-y^{(k)})\sigma_{N-k+2} \\
&+ (y^{(k-2)}-y^{(k-1)})\sigma_{N-k+3}+ \dots\\
&+ (y^{(1)}-y^{(2)})\sigma_{N} \Big)\Big|
\end{rcases}\text{$\sigma_j \le \delta$ \qquad\qquad\quad     }
\end{flalign*}


which we can rearrange according to \(y^{(j)}\) instead to
\begin{equation*}
\begin{aligned}
e^{(k)}= & \Big| y^{(k)} - \Big(\Sigma_{j=k+1}^{N} y^{(j)} [\sigma_{N-j+1}-\sigma_{N-j+2}] \\
& + y^{(k)} [\sigma_{N-k+1}-\sigma_{N-k+2}]  + \Sigma_{j=2}^{k-1} y^{(j)} [\sigma_{N-j+1}-\sigma_{N-j+2}] + y^{(1)}\sigma_N \Big) \Big|
\end{aligned}
\end{equation*}

Rewriting \(d_i=\sigma_{i+1}-\sigma_i\), we now have
\begin{equation*}
\begin{aligned}
e^{(k)}= & \Big| \Sigma_{j=k+1}^{N} y^{(j)} d_{N-j+1} + y^{(k)} [1-(1-\delta - \delta)] + \Sigma_{j=2}^{k-1} y^{(j)} d_{N-j+1}+ y^{(1)}\sigma_N \Big) \Big|
\end{aligned}
\end{equation*}

As either \(\sigma_i \ge 1-\delta\) or \(\sigma_i \le \delta\), we have \(|d_i|\le\delta\) for all applicable \(i\), and using triangle inequality,
\begin{equation*}
\begin{aligned}
e^{(k)}\le & |y^{(k)}|\delta+ \Sigma_{j=1}^{N} |y^{(j)}|\delta \le \delta (N+1) \max_{k}{|y^{(k)}|}
\end{aligned}
\end{equation*}
and we are done. \(\square\)

\subsubsection{TNN error bound and resistance to catastrophic forgetting}

\textbf{Proposition \ref{prop:errorbound}}. \textbf{TNN does not suffer from CF}. Let fitting and external datasets be \(D,D'\). Let the user specified tolerance be \(\epsilon\). There exists \(A\subseteq D'\) such that if TNN is constructed with \(D\cup A\), then for any \((x,y)\in D\cup D'\), sample-wise error \(e=|y-TNN(x)|<\epsilon\).  \label{proof:errorbound}

Proof: Let \(A=\{(x',y')\in D':e=|y'-TNN(x')|\ge \epsilon\). Construct new TNN with \(A\cup D\) and the same tolerance \(\epsilon\), then, for any \((x',y')\in A\), we obtain \(e<\epsilon\). 

This does not immediately guarantee the error bound. This is best illustrated in fig. \ref{fig:errorupperbound}(B). However, the procedure above can be repeated. Suppose \(A'=\{(x',y')\in D'-A:e=|y'-TNN(x')|\ge \epsilon\}\), then let \(A\rightarrow A\cup A'\). Again, for any \((x',y')\in A\), we obtain \(e<\epsilon\). Repeat this until the desired condition is achieved or \(A=D'\). 

TNN is resistant to catastrophic forgetting for clear reasons: each data sample in \(D\) still corresponds to one exact neuron in TNN during each reconstruction process when \(A\) grows in size. \(\square\)

\begin{figure}[htbp]
\centering
\includegraphics[width=0.8\textwidth]{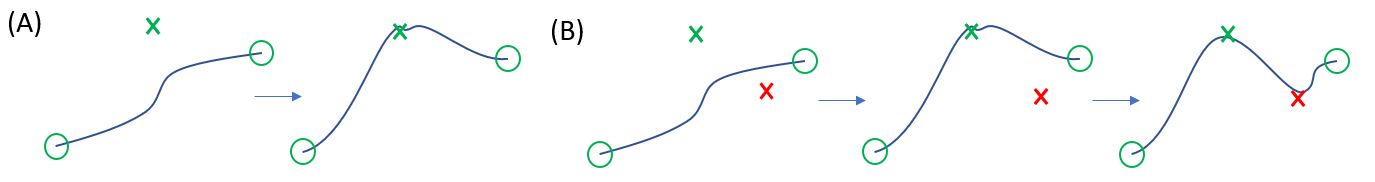}
\caption{Green circles: fitting data samples used in TNN. Green and red x: external samples. (A) The inclusion a novel or unseen external sample into the fitting dataset (B) The inclusion of one external sample (green x) causes another external sample (red x) to be out of distribution, and thus it needs to be included in the next iteration.} 
\label{fig:errorupperbound}
\end{figure}

\subsubsection{TNN example}
\label{TNNexample}
\textbf{TNN pencil-and-paper example}. Use TNN to fit the dataset  \((x,y)\in\{(1,1),(0.5,2),(0,3)\}\). Then \(f(x)=3\sigma(20x+5)-\sigma(20x-5)-\sigma(20x-15)\). 
Suppose \(a=5\) and we have a dataset \(\{(1,1),(0.5,2),(0,3)\}\) so that \(N=3\) and \(x^{(1)}=1>x^{(2)}=0.5>x^{(3)}=0\), which is evenly spaced thus we can use the simplified formula, for example, \(W_k=2\times 5\times (3-1)=20\) for \(k=1,2,3\) and \(b_1=5(3-2\times 1)\) etc. Then \(W=[20,20,20]^T\), \(b=[5,-5,-15]^T\) and 
\begin{equation*}
\begin{aligned}
\alpha & = A^{-1}y= \begin{pmatrix} 0 & 0 & 1 \\ 0 & 1 & -1 \\ 1 & -1 & 0  \end{pmatrix} \begin{bmatrix} 1 \\ 2 \\3 \end{bmatrix} = \begin{bmatrix} 3 \\ -1 \\ -1 \end{bmatrix}\\
f(x) & = \alpha^T\sigma(Wx+b) = [3,-1,-1]\sigma([20,20,20]^T x + [5,-5,-15]^T)
\end{aligned}
\end{equation*}
where \(\sigma\) is applied component-wise, thus \(f(x)=3\sigma(20x+5)-\sigma(20x-5)-\sigma(20x-15)\). 

\subsubsection{More remarks}
There is also a mid-point property that can be exploited for generalizability to arbitrarily high accuracy, where data must be sampled such that any instance \(x_{external}\) lies inside either (1) the fitting dataset or (2) is equal to some mid-point of two neighbouring fitting samples; see the proposition below. Fig. \ref{triangularcon}(B) shows how the component of \(x_{mid,k}\) at \(j=N-k+1\) is half-activated i.e. the activation value is \(0.5\).  Admittedly, this is an ideal condition for accurate generalizability.

\label{proof:midpoint}
\textbf{Mid-point property}. The mid-point \(x_{mid,k}=\frac{1}{2}(x^{(k)}+x^{(k+1)})\) takes the value of \(\alpha^T\sigma(Wx_{mid,k}+b)=\frac{1}{2}(y^{(k)}+y^{(k+1)})\). 
Proof: \(\sigma_j^{mid}\equiv\sigma(Wx_{mid,k}+b)_j=\sigma(\frac{1}{2}(Wx^{(k)}+b)+\frac{1}{2}(Wx^{(k+1)}+b))_j\). Then \(\sigma_{N-k+1}^{mid}=\sigma(\frac{1}{2}a+\frac{1}{2}(-a))=\sigma(0)=0.5\). For \(j\le N-k\), \(\sigma_j^{mid}=1\), while for \(j\ge N-k+2\), \(\sigma_j^{mid}=0\). The resulting output of the neural network looks like \(\alpha^T[\dots,1,0.5,0,\dots]^T\), which is equal to \(\frac{1}{2}\alpha^T[\dots,1,1,0,\dots]^T+\frac{1}{2}\alpha^T[\dots,1,0,0,\dots]^T=\frac{1}{2}(y^{(k)}+y^{(k+1)})\) \(\square\).

\label{tnnremarks}
\textbf{Smoothness}. From the construction, assuming sigmoid function as the activation function, it is obvious that the function is continuous for finite \(a\). As \(a\) increases, the function becomes more and more constant around each data sample as shown in fig. \ref{fig:3exps}(A1-3), i.e. becoming more step-wise. 

\textbf{Special case}. When the dataset is evenly spaced, \(x^{(k)}=1-(k-1)\Delta\), \(k=1,...,N, \Delta=1/(N-1)\), the results simplify to \(W_k=2a(N-1)\) and \(b_k=a(3-2k)\). Not only equation (\ref{maincondition2}) is fulfilled, we also get \((Wx^{(k)}+b)_j = a(1+2[N-k+1-j])\). For the k-th data sample, the activation will then be well-spaced in an interval of \(2a\), so that \(\sigma^{(k)}=\sigma([\dots,-3a,-a,a,3a,\dots])^T\approx [\dots,0,0,1,1,\dots]^T\).

\textbf{Scalability and complexity}. The bulk of memory space usage comes from \(W\in\mathbb{R}^{N\times n}\). \(N\) is the number of available data points, and this is an unusual feature compared to modern DNN architecture. Since \(N\) in common datasets can grow very large, the space complexity becomes \(\Omega(Nn+mn)\), still linear w.r.t \(N\). Since it is more likely that \(m<N\), it is reasonable to simplify to just \(\Omega(Nn)\). To reduce the complexity, we can pick a set of representative data points \(X_{rep}=\{x_{rep}\}\) for NN construction. The selection of representatives depends on our error tolerance, and it can be done through other machine learning methods, such as clustering. Other data points can then be used for validation. The resulting complexity \(\Omega(N_{rep}n)\) will thus highly depend on the variability and the structure of the dataset where \(N_{rep}=|X_{rep}|\). Time complexity is almost irrelevant for now, since we are not able to find any meaningful way to compare with the training process of modern DNN. As far as we know, there is no decisive rule on how many epochs are necessary for a DNN training through back-propagation.

\textbf{Generalizability to n-dimensional output}. Generalization to scalar input and multi-dimensional output is relatively simple. From equation (\ref{eq:form}), we can treat \(\alpha\) as the coefficients for the only component of one-dimensional \(y\). Generalizing to \(y\in\mathbb{R}^m,m>1\), identify each vector \(\alpha_i\) with the component \(y_i\). Stacking them up, we can redefine \(\alpha=[\alpha_1^T; \alpha_2^T; \dots]\) where a semi-colon denotes the next row, and the construction is done. Note that now \(\alpha\in\mathbb{R}^{m\times N}\).

Further generalization to n-dimensional input has been unsatisfactory (see later section of appendix, \ref{extra:TNNgen}). Instead, SQANN has been developed with additional mechanism that TNN does not possess.

\subsection{Appendix for SQANN}
\label{appendix:sqann}
\textbf{Ill-defined dataset}. We exclude any dataset which is ill-defined, i.e. when there exists two identical \(x^{(k)}=x^{(k')}\) having different \(y^{(k)}\neq y^{(k')}\). Such dataset is not suitable for any function and will cause unresolvable collisions in our models. 

Link back to main text section \ref{section:SQANN}.

\subsubsection{Selective Clustering}
\label{appendix:selectiveclustering}
\textbf{Selective clustering} has been loosely defined in the main text. They are addressed here. Formally and \textit{more generally}, selective clustering is defined as the following. Let \(A=X\times Y\) be a set, and \((x,y)\) be a point. Define \(d(A,x)\) as the minimum distance between \(x\) and all points in \(X\), \(x_1=argmin_{x'\in X}|x-x'|\). Then let \((x_1,y_1)\in A\). Let error tolerances be \(\delta,\epsilon>0\). Suppose \(d(X,x)\le\delta\), then (case 1) if \textit{output values are similar} \(|y-y_1|\le\epsilon\), then add \((x,y)\) as a new member of \(A\). Otherwise (case 2), if \textit{output values are distinct}, \(|y-y_1|>\epsilon\), then let \((x,y)\) form its own new cluster. In short, a new cluster is created (at least nominally) if we have neighbours with distinct \(y\) values; they are neighbours, but we may be looking at points at different sides of classification boundaries. We thus need a function to decide which eventual value is taken. In this paper only argmax of neuron activations is considered, and the value related to the cluster with the maximally activated neuron is returned. Thus we have a selective clustering algorithm. \textit{Remark}. This can be easily generalized such that \((x,y)\) is added to another set \(B\) in case 1 where \(|y-y_2|<\epsilon\) for \((x_2,y_2)\in B\), or form another cluster otherwise in case 2. 

As mentioned in the main text, this does have a role for interpolation during SQANN propagation. Suppose \(x_z\) is near both \(x\) and \(X\). In our models, output \(y_z\) is determined by taking argmax to find the neuron that is strongly activated (only if it exists). When no neuron is activated strongly enough, interpolation is performed. Selective clustering is the mechanism that handles this. In case 1, i.e. when \(y,y_1\) are similar, the approximation of \((x_z,y_z)\) is likely good because the output of \(x_z\) computed from taking interpolation between strongly activated neurons within \(X\) (including \(x\)) will yield approximately similar to the value obtained by taking argmax (the most strongly activated neuron). This is the good case. 

However, in case 2, we can get unstable result. This is because \(x_z\) might be near enough to both \(x,x_1\) that two different values \(y,y_1\) appear to be equally correct. Interpolation on \(y,y_1\) could yield the ``averaged" value which may not reside in either cluster, which might cause even more uncertainty. The concept selective clustering is thus introduced with the express purpose of resolving such situation. During collision, if two samples strongly activate the neurons but have distinct values, we must be careful about how we decide the output (argmax or interpolation). In this paper, only argmax is used i.e. the boundary between selective clusters are crisp. For example, when \(x_z\) causes activation of 0.99 on one neuron and 0.92 on another. Both are strong activations, but with argmax-based selective clustering, \(x_z\) is treated as a member of the first neuron's strong activation cluster, thus the \(y\) attached to the first neuron will be used as the output. This helps the models achieve UA on \(D\). Validation on external dataset shows reasonable accuracy achieved too.  

There might not exist a best universal choice to handle such boundary cases, considering that the shape of local manifold may change depending on the dataset distribution. Hence, we leave other variations for further studies.

\begin{figure}
\centering
\includegraphics[width=0.6\textwidth , ]{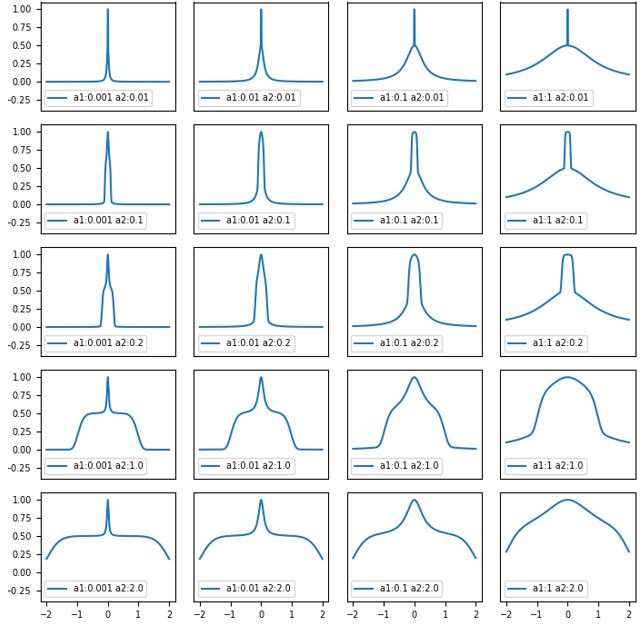}
\caption{More \(a_1,a_2\) variations of double selective activation.} 
\label{fig:sigmadsa}
\end{figure}

\subsubsection{Computing output via SQANN propagation (prediction)}
\label{appendix:sqannprop}
Fig. \ref{fig:sqannprop} shows two cases occurring during SQANN propagation. Case (A) shows a case when strong activation occurs at layer 2 node 3, i.e. \([v_2]_3>\tau_{act}\), near 1 or even exactly 1 (if \(x\) is the exact fitting sample used during construction). Then \(y=[\alpha_2]_3\). In case (B), no nodes are strongly activated. The \(\alpha\) values of the two most strongly activated neurons are used for weighted average based on the strength of activations \([v_2]_3\) and \([v_3]_2\), i.e. interpolation is performed. Clearly, we can explore different variations, for example, taking three most strongly activated neurons etc. 

This is a link back to main text section \ref{section:SQANN}.

\begin{figure}
\centering
\includegraphics[width=0.8\textwidth , ]{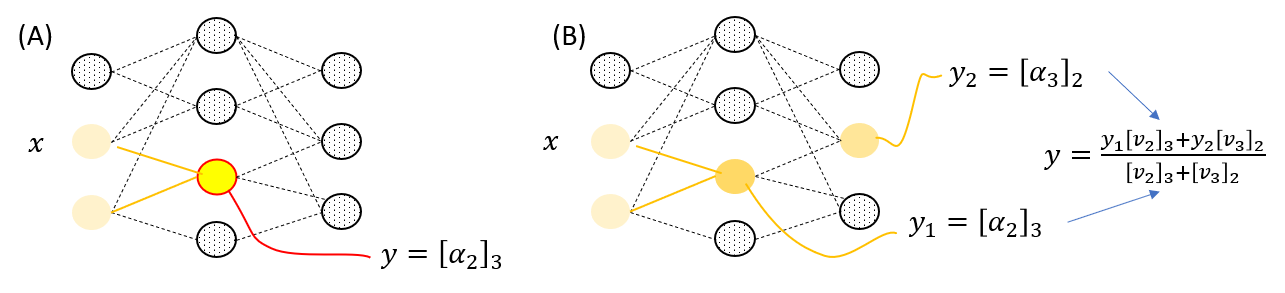}
\caption{(A) Strong activation at layer 2 node 3 (yellow circle with red boundary). The output is taken as \([\alpha_2]_3\) (B) No strong activation, only 2 moderate activations (circles with darker shades of oranges). The output shown is a weighted average.} 
\label{fig:sqannprop}
\end{figure}

\subsubsection{Good practice for scalability}
\label{appendix:scalability} 
For both TNN and SQANN, if there are 1 million fitting data samples, there will be 1 million neurons. Do we need all 1 million data? Probably not, since it might be fair to assume that some data points in the fitting dataset may have similarities. This is where pre-processing can be done to ideally root out data that are redundant. SQANN itself can be used to check how redundant they are; for example, a subset of the data can be collected and a smaller SQANN can be constructed for testing purposes. We can check how the subset of `similar data' activates each other's neuron within this mini SQANN.

This is a link back to \textit{computing output via SQANN}: go to lemma \ref{lemma:first}.

\subsubsection{SQANN construction can complete}

\label{appendix:sqanncompleteproof}
\textbf{Definitions}. Let us start with some terminologies and sketch of proof before diving into the assumption required for SQANN to complete.
\begin{enumerate}[leftmargin=*,topsep=0pt]
\item Let the \textbf{activation space} \(\mathcal{A}\) be a high dimensional space of pattern activations. An element of \(a_{lk}\in\mathcal{A}\) is a collection of \([v_l]_k\) obtained by collecting the signals of neurons during SQANN propagation of a sample \(x\). 
\item \textbf{Distance and strength of activations}. The distance between any two \(a_{lk},a_{l'k'}\) can be chosen arbitrarily e.g. we can use standard Euclidean distance. Fig. \ref{fig:sketchofcomplete}(B) illustrates the activation \(a\) of a sample (red x mark) and red/blue/green plus marks are activation of other samples that strongly/intermediately/weakly activate neuron corresponding to red x (whose activation is \(a\)). In SQANN construction, readers might have noticed that the ``distance" is related to eq. (\ref{eqn:act}). Let \(x\) be an input. Let \((x',y')\) be a sample admitted to SQANN node \(j\) of layer \(k\). Then the distance of activations between \(x\) and \(x'\) is precisely the \(||.||\) term of eq. (\ref{eqn:act}) if \(v\) is the activation signals of \(x\) propagated to layer \(k\).

\item  Let any sample \(u=(x^{(u)},y^{(u)})\in U\) where \(U\subset D\) is the subset of fitting dataset not yet used during SQANN construction, \(s\in S\) where \(S=D-U\) is the subset whose samples are already integrated as nodes of SQANN. Let \(v^{(u)}\) the synapse of \(u\) at layer \(l\). Let \([v^{(u)}_m]_n\le\tau_{act}\) on all \(m,n\), where each \(m,n\) corresponds to an existing node \(\eta_s\) constructed from \(s\in S\) i.e. let \(u\) not activate any existing nodes too strongly. \textbf{Complication} is defined as the following phenomenon: \(s_1\in S\) activates some \(s\) strongly, but \(u\) also activates \(s_1\) strongly. We then define the \textbf{probability of complication}, i.e. probability that a complication occurs for \(u\in U\) to be \(p^{c}_u\). 

\item  Let \(A(x)\) be the set of points strongly activating \(x\) (in the illustration, this is the area covered by the small red circle).  Also let \(d\) (illustrated in fig. \ref{fig:sketchofcomplete}(B)) denote the minimum distance between \(u\in U\) and any points within \(A(s)\) for any \(s\in S\). 

\item Dataset is \textbf{\(p^{c}_u\)-sparse} if for a given \(\tau_{act},\tau_{ad}\) and the current ordering of dataset, there exists \(d>0\), such that for any \(u\in U\), for any \(s\in S\), if the minimum distance between \(A(s)\) and \(u\) is greater than \(d\), then the probability of complication \(u\in A(s_1)\) for any \(s_1\in S\) is at most \(p^{c}_u\).

\item Furthermore, ideally, each fitting sample is expected to represent its locality well so that no sample is redundant. A redundant sample can be illustrated as the following. Suppose \(u\) perfectly captures the locality \(A(u)\), i.e. any data \(s\) causes strong activation of the node \(\eta_u\) in SQANN if and only if \(s\in A(u)\).   A data sample \(u'\) is \textbf{redundant} if \(A(u')\subseteq A(U)\) and \(u'\) comes after \(u\) in the ordered dataset. A fitting dataset is \(n\)-redundant if it contains \(n\) redundant data points.

\end{enumerate}

\textbf{Sketch of proof}. The following is a series of descriptions and intuitions leading to the proof for proposition \ref{prop:complete}.

The main concern about the completion of SQANN construction mainly stems from the collision problem i.e. if \(a'\) is like the red + mark. During collision, layers are torn down, and the red + mark is now added beside the red x mark; thus SQANN now can distinguish similar looking points that might be characterized differently (\textit{selective clustering} in action). Let us generically refer to the activation of some sample \((x_o,y_o)\) that is already integrated into SQANN layers as \(a_o\) (represented by black open circles). The question is, will \(a_o\) be reconstructed back to the previously destroyed layers in the same manner as they were?  The answer is, they will NOT be exactly the same as before since the new activations now has one more dimension at layer \(k\) thanks to the new node contributed by \(a'\). More importantly, will they cause collisions that did not occur before? We need an answer to this question if we were to prove that SQANN construction can complete. 

Unfortunately, it might not be possible to answer that question in few sentences. Instead, from here on, we will go through the intuitions depicted in fig. \ref{fig:sketchofcomplete} and define ``complication". Once we understand what a complication is, we will be able to talk about the rate at which recurring collisions occur, hence the probability of completion. 

\textit{Intuition behind \(\sigma_{dsa}\), double selective activation}. We want to find the condition in which there is a high probability that all \((x_o,y_o)\) that were removed from the layers during collision resolution mechanism will come back in the same order. This is the exact reason why double selective activation has a narrow band of strong activation (small area encircled by red band in fig. \ref{fig:sketchofcomplete}) distinguished from a band of moderate activation (area encircled by orange band) by steep gradient, that is in turn distinguished from near zero activation (area outside orange band) by steep gradients. In fig. \ref{fig:sigmadsa}, they are respectively (1) the spiky area near zero, (2) the intermediate values to the left and right of the spike and (3) any other typically low magnitude values beyond. 

With such characteristic, \(\mathcal{A}\) will have a small red region of strong activation, a large orange buffer region of moderate activation and weak activation everywhere else. The main idea revolves around the fact that the buffer region reduces the probability that existing black open circles are too close to red + mark. Hence, the activation characteristic of each black open circle that has been temporarily removed from the torn down layer (because of collision) will be similar to the activation characteristic it has before i.e. it will not suddenly strongly activate a node it previously did not activate. To spell this out more clearly with an example, suppose sample \(u\) weakly activates all existing neurons with values 0.01 before. After red + mark is added into the layer, \(u\) still weakly activates all existing neurons with values 0.01, including red + mark which is close to red x mark. More precisely, let the activation of \(u\) is given by \(v\) before collision. Let us denote its component at layer \(l=1,2,\dots,l_c,\dots,L\) by \([v_l]_i,i=1,\dots,n_l\)). Then after resolving collision, the new activation is \(v'\) such that \([v_l']_i\approx [v_l]_i,i=1,\dots,n_l\), although now \([v_{l_c}']\) has the \((n_{l_c}+1)\)-th component since the collided sample has been added into layer \(l_c\).

\begin{figure}
\centering
\includegraphics[width=0.9\textwidth , ]{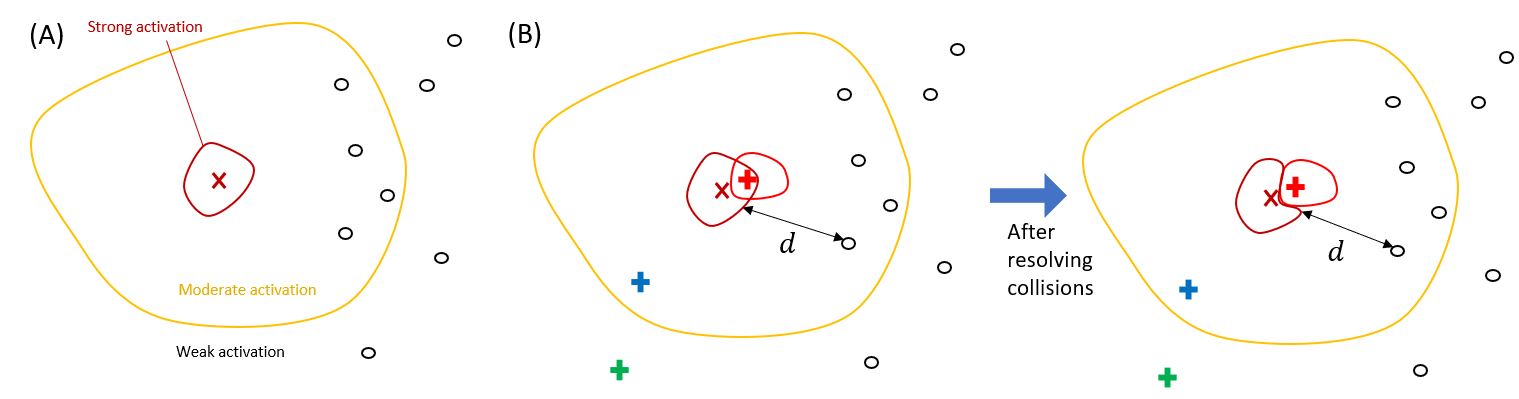}
\caption{(A) Visualization of a SQANN node \(\eta\) in the \textit{activation space}, marked as red x. Black open circles are the fingerprints of other nodes that are already integrated into SQANN. (B) A data sample whose activation lies within the red/orange region such as red/blue plus mark is strongly/moderately activating the particular node \(\eta\). Otherwise, it is weakly activating it, e.g. green plus mark. A dataset and SQANN construction settings are desirable if for every fitting data \(u\), there is a reasonably large \(d\) such that the probability that complication occurs is \(p_u^c\approx 0\). When a data sample \(x'\) strongly activates node \(\eta\) (red plus mark), it will cause collision. If this happens, \(x'\) will be integrated into the collision layer as \(\eta'\), i.e. collision is resolved, hence shifting the \textit{effective} shape of activation shape of \(\eta\) shape (blue arrow). What used to strongly excite \(\eta\) might now excite \(\eta'\) more strongly (say, if it is nearer to \(\eta'\)).} 
\label{fig:sketchofcomplete}
\end{figure}

\textit{The illustration of an iteration in SQANN construction}. Let red x in fig. \ref{fig:sketchofcomplete} represent activation \(a_{lk}\) in \(\mathcal{A}\). Let it be the activation of a fitting sample \((x^{(m)},y^{(m)})\in D\); in other words, sample \(m\) has been integrated into layer \(l\) as node \(k\). Suppose the fitting sample \((x',y')\in D\) is used in an iteration of SQANN construction and its activation at layer \(l\) is \(a'\). We perform \textit{check \(N_k\) activation} on the sample: from the main text, we see that 3 possibilities can occur. (1) Admission of sample to layer \(N_k\), if the sample weakly activates the existing node without collision, e.g. \(a'\) is represented by the green + mark in fig. \ref{fig:sketchofcomplete}(B) (2) collision if it activates an existing node very strongly, e.g. \(a'\) is like the red + mark (3) filtering into deeper layer if neither occurs, e.g. \(a'\) is like the blue + mark.

\textit{Intuition on well-behaved iteration in SQANN construction}. Moving on from the illustration, consider the following: if a fitting sample \(u\) does not activate any nodes strongly and another fitting sample \(s_1\) activates an existing node corresponding to a fitting sample \(s\) strongly, then \(u\) will not activate the node constructed from \(s_1\) strongly. Intuitively, this is because \(s_1\) is near \(s\), which is far from \(u\) because there is a buffer region, thus \textit{there is a high probability that \(u\) is also far from \(s_1\)} (note: by \textit{far}, we mean distance in the activation space). However, this is not a universal mathematical property. It is only intuitive that similar samples are similar in activation space too, and we assume that for practical dataset, this assumption holds with high probability, i.e. we allow the possibility that of pathological shape of buffer where ``unlike" samples activate each other strongly. We formalize this as the following \textit{probability of complication}.

\textit{Further considerations}. In the case where there are too many redundant samples, what we really need to consider are (1) our selection of \(\tau_{ad},\tau_{act}\) might be unsuitable, since they cause too many overlaps. As a rule of thumb, decrease \(\tau_{ad}\) and increase \(\tau_{act}\) to enable constructions with less overlaps between samples' activations (2) there are too many similar data in the samples that could have been represented equally well with a fraction of available data. In this case, it might be better to remove some data from the samples and create smaller subsets for SQANN layer construction, which is beneficial, since the remaining data can be used for validations. However, in this paper, we do not perform any procedure to filter redundant data yet. This will be left for future study, and we focus on integrating ALL \textit{fitting data} for UA.

\textbf{Proposition \ref{prop:complete}}. If \(D\) is \(p^{c}_u\)-sparse and \(n\)-redundant, then SQANN construction completes with at least \((1-p^{c}_u)^{n/|D|}\) probability. If there is no complication, SQANN construction completes.

Proof. The first statement directly follows from the definitions. The second statement means using a stronger assumption \(p^{c}_u=0\) i.e. assume there is no complication at each sample checking step. Suppose layer \(1,\dots,k-1\) have been constructed, and \(j-1\) nodes have been added to the latest layer \(k\) that is being constructed. Recall and note the difference between \(x^{(k)}\) and \(x^{<k>}\). Suppose  sample \(x^{(j)}\) causes collision in layer \(c\le k-1\). Then we tear down all layers after layer \(c\); here is where we will use the \textit{order integrity}: let \(X_{temp}=\{x^{<r>},x^{<r+1>}\dots,x^{<j-1>}\}\) be the set of samples that have been returned to the list of unused indices in the same order they have been put into SQANN layer (likewise \(Y_{temp}\)) e.g. \((x^{<r>},y^{<r>})\) is the first sample in layer \(c+1\). Then, \textit{resolve the collision} by concatenating \((x^{(j)},y^{(j)})\) to \((N_c,\alpha_c)\) of layer \(c\).  The reconstruction of subsequent layers will occur as the following. Check \(x^{<r>}\) for admission, add it into layer \(c+1\), then checking \(x^{<r+1>}\) the same way it was added through ``layer \(k\) construction" process in the main text, and so on up to \(x^{<j-1>}\). They will be returned to their previous positions the same way they were added to the torn down layers. By the no-complication assumption, samples will not collide with \((x^{(j)},y^{(j)})\). More verbosely, samples admitted via ``admission to \(N_k\)" will again be admitted to the same \(N_k\) at the same node, while sample admitted through collision with some node \(\eta_{k'}^{<j'>}\) will collide with the same node, and \textit{there is no collision with the node of activation corresponding to} \((x^{(j)},y^{(j)})\) because of no complication assumption. As a consequence, we can always proceed with \(x^{(j+1)}\), even if there is a need for multiple reconstructions of layers. Thus, the construction can complete. \(\square\)

\subsubsection{SQANN Main Theorem}
\label{proof:theoremsqann}
\textbf{Theorem \ref{theorem:sqann}}. Assume SQANN construction is completed. SQANN is a UA on \(D\). Furthermore, it is resistant to CF. 

Proof: Let the fitting dataset be \(D=X\times Y\). For any \(x\in X\), perform SQANN propagation. If \(x\in N_1\), then, by lemma \ref{lemma:first}, we obtain the arbitrarily high accuracy. If \(x\in N_k,k>1\), the heavy-lifting is in fact already done through the SQANN construction algorithm. By SQANN construction, there exist index \(j\) and layer \(k>1\) such that either (1) the activation of \(x\) undergoes \textit{admission to \(N_k\)} as \(\eta_k^{<j>}\) or (2) pushed into layer \(N_k\) during collision. One of the two must occur, otherwise SQANN construction is not completed, contradicting the assumption. In either process, we have \(v_{k-1}=\eta_k^{<j>}\) thus \([v_k]_j=\sigma_{dsa}(|| v_{k-1}-\eta_k^{<j>}||)=\sigma_{dsa}(0)=1>\tau_{act}\) where recursive computation \([v_l]_{i}=\sigma_{dsa}\big(||v_{l-1}-\eta_{l}^{<i>} ||\big)\) for all \(i=1,\dots,n_l\) is performed from layer \(l=1,\dots,k\) with \(v_0\equiv x\). Since such strongly activated neuron exists at layer \(k\), by SQANN propagation, retrieve \(y=\alpha_k^{<j>}\) where \(j=argmax_{j'}[v_k]_{j'}\). To ensure that \(j\) is unique, we reasonably assume that there is no duplicate \(x\) with different \(y\) values that is admitted through collision, otherwise the dataset is ill-defined (as previously mentioned). The uniqueness is made possible because double selective activation is, by definition, has a peak with unique value 1 and in the locality of the peak, to each side, it is one-to-one and onto, i.e. there is no interval of constant value around the peak. Finally, \(j\) is indeed the index from argmax, because the peak value of double selective activation is 1 by design.  

\textit{Resistance to catastrophic-forgetting}. For each new fitting sample added into SQANN, previous samples still exist as nodes stored in the SQANN layers. Previously learned samples will therefore not be forgotten. More precisely, suppose SQANN construction is not yet complete. Let \(D_k\subset D\) be the subset of dataset whose samples have either been admitted to layers \(1,2,\dots,k\) through normal admission or by resolving collision. Then for \(p_k=(x^{(k)},y^{(k)})\in D_k\) can be exactly queried by obtaining activation at some layer \(l\le k\)  for some \(j\) so that \(v=\sigma_{dsa}(||v_{l-1}- \eta_l^{<j>}||)\) is exactly \(1\) and \(l,j\) exist exactly at the layer and node where \(p_k\) is admitted into SQANN through the main mechanism of layer construction. By SQANN propagation, as before, we retrieve the output \(\alpha_l^{<j>}=y^{k}\) stored exactly at \(l,j\) node through the main SQANN construction algorithm as well, hence, SQANN remembers the previously stored value. Furthermore, if SQANN construction with \(D\) has been completed and new fitting dataset \(D'\) is available, construction with all samples in \(D\) by drawing samples in the same order will result in the same SQANN. Any new samples from \(D'\) can be admitted into SQANN in the same way specified in \textit{layer \(k\) reconstruction} \textit{after} all \(D\) samples are used up. In this manner, previously learned samples from \(D\) will remain inside SQANN layers, i.e. like before we can find \(l,j\) node corresponding to each sample previously admitted into SQANN. Hence no forgetting will occur. \(\square\) 

\subsubsection{SQANN example}
\label{appendix:sqannexample}
\textbf{SQANN pencil-and-paper example}. With \(a_1,a_2=0.001,0.5\), \(\tau_{ad},\tau_{act}=0.1,0.9\), create SQANN universal approximator for indexed data \(X=[x^{(1)},x^{(2)},x^{(3)},x^{(4)}]=\big[\begin{smallmatrix}
1 & 1.2 & -1 & -1.2\\1.2 & 0.8 & -1 & -1.2\end{smallmatrix}\big]\) and \(Y=[y_1,y_2,y_3,y_4]=[1,1,0,0]\). (A) Show layer 1 stores the fingerprints of \((x^{(1)},x^{(3)})\) and layer 2 stores \((v^{(2)},v^{(4)})\) i.e. activations of \((x^{(2)},x^{(4)})\). (B) Use SQANN propagation to verify that we indeed get zero errors on \(X\times Y\). (C) Test SQANN on the external dataset \(X_{ex}=[x_t^{(1)},x_t^{(2)},x_t^{(3)}]\big[\begin{smallmatrix}
1.25 & -1.25 & -1 \\1.25 & -1 & -1.4\end{smallmatrix}\big]\) and plot the results, marking the interpolations made by SQANN.

Note that the following can be matched with the demonstration in jupyter notebook SQANN\_small\_example.ipynb.

(A) \textit{SQANN construction}. We start by putting \(x_1,y_1\) into the first layer, so put it into the first layer of SQANN, \(N_1=(x_1),\alpha_1=(y_1)\). Now check \((x^{(2)},y^{(2)})\) for admission to \(N_1\): if it activates a node in \(N_1\), then we \textit{filter it to a deeper layer}, otherwise, we add it into \(N_1,\alpha_1\) as well. We show that the latter occurs. We only have one node in SQANN now, so the only possible activation is \([v_1^{(2)}]_1=\sigma_{dsa}(||x^{(2)} - x^{(1)} ||)=0.3344\ge\tau_{ad}\). Admission to \(N_1\) only occurs if \([v_1^{(2)}]_1<\tau_{ad}\) hence it is filtered to a deeper layer. 

Now we check \((x^{(3)},y^{(3)})\) for admission to \(N_1\) and get \([v_1^{(3)}]_1=5.655\times 10^{-5}\). It does not activate any node in the layer, thus this is a distinct sample we will admit into \(N_1\). 

Now we check \((x^{(4)},y^{(4)})\) for admission to \(N_1\). We have two nodes, so we have to compute both: \([v_1^{(4)}]_1=\sigma_{dsa}(||x^{(4)} - \eta_1^{<1>} ||)=4.450\times 10^{-6}\) and \([v_1^{(4)}]_2=\sigma_{dsa}(||x^{(4)} - \eta_1^{<2>} ||)=0.5676>\tau_{ad}\). It does activate a node in \(N_1\), which is \(\eta_1^{<2>}\). Hence it is filtered to a deeper layer. We have shown that \(N_1\) stores \((\eta_1^{<1>}=x^{(1)},\eta_1^{<2>}=x^{(3)})\), and not the other samples. Note: since the activation \(<\tau_{act}\), no node has been activated \textit{strongly} (otherwise we will have collision). Recall that each time we store \(x^{(k)}\) into \(N_l\), we also store \(y^{(k)}\) into \(\alpha_l\), so now \(\alpha_1=(y^{(1)},y^{(3)})\).

We have gone through the fitting dataset once. The unused data are \(\{(x^{(k)},y^{(k)}), k=2,4\}\), as they have been filtered to deeper layer. We now proceed with layer 2 construction. Since it is empty, we put the activation of \(x^{(2)}\) (not the sample itself) into \(N_1\) since no collision occurs, i.e. \(\eta_2^{<1>}=v_2^{(2)}=\big[ [v_2^{(2)}]_1, [v_2^{(2)}]_2\big]\). To show no collision, i.e. no strong activations in \(N_1\), similar to before, we compute \([v_1^{(2)}]_1<\tau_{act}\), which is previously done, and \([v_1^{(2)}]_2= 6.187\times 10^{-5}<\tau_{act}\).

The last sample \(x^{(4)}\) is admitted into \(N_2\) by checking collisions against all \(N_1\) (previously done), and then checking activation against existing \(N_2\) node, \(\eta_2^{<1>}\), i.e \(v_2^{(4)}=\sigma_{dsa}(|| v_2^{(4)}- \eta_2^{<1>}||)=0.00732<\tau_{ad}\). Hence, it is admitted to \(N_2\). We have used up all data points, hence we have shown \(N_2=(v_2^{(2)},v_2^{(4)})\).

(B) Using SQANN propagation on \(x^{(k)},k=1,3\), we get \(\sigma_{dsa}(|| v_1^{(k)}- \eta_1^{<j>}||)=\sigma_{dsa}(0)=1>\tau_{act}\) where \(j=1,2\) respectively. Since they are strongly activated, we get \(y=\alpha_1^{<j>}=y^{(1)},y^{(3)}\) for \(j=1,2\) respectively. Thus the errors are zero.

For \(x^{(2)}\), we already previously computed \(v_2^{(2)}=\big[ [v_2^{(2)}]_1, [v_2^{(2)}]_2\big]\), which becomes \(\eta_2^{<1>}\) thus we will also get the distance of activation value from itself as stored in \(N_2\), \(\sigma_{dsa}(||v_2^{(2)}-\eta_2^{<1>}||)=\sigma_{dsa}(0)=1\). Likewise \(x^{(4)}\).

(C) \textit{SQANN testing}. For \(x_t^{(1)}\), we expect its fingerprint to be close to \(x^{(1)}\) or the activation of \(x^{(2)}\) since their values are similar. It turns out we get the following activations in layer 1, \(v_{t,1}^{(1)}=[0.5053, 4.938\times 10^{-5}]\), not strongly activating layer 1 nodes. For layer 2, we have\([v_{t,2}^{(1)}]_1=\sigma_{dsa}(||v_{t,1}-v_2^{(2)}||)=0.5165\) and \([v_{t,2}^{(1)}]_2=0.0009896\). There is no strong activation anywhere, and this is the case when interpolation is needed. Notice that of all the activations we computed, the strongest are \([v_{t,1}^{(1)}]_1\) and \([v_{t,2}^{(1)}]_1\), which are the first nodes in both layer, due to \(x^{(1)}\) and  \(x^{(2)}\) respectively as we have expected. Using SQANN propagation, we fetch its \(\alpha\) values, both of which are \(1.\), thus using linear interpolation, \(y=\frac{0.5053*1+0.5165*1}{0.5053+0.5165}=1\).

The next sample in the external dataset \(x_t^{(2)}\) does activate the SQANN. We expect it to activate either \(x^{(3)}\) or the activation of \(x^{(4)}\). First, get \(v_{t,1}^{(2)}=[5.049\times 10^{-5},0.5059]\), i.e. no strong activation, so \(x^{(3)}\) is not strongly activated. But we get \(\sigma_{dsa}(||v_{t,1}^{(2)}-\eta_2^{<2>}||)=0.9880>\tau_{act}\), thus we do get strong activation due to the fingerprint of \(x^{(4)}\). By SQANN propagation, we get \(y=\alpha_2^{<2>}=y^{(4)}=0\). 

The last external sample needs interpolation too, and we leave it for the reader. The plot of results are shown in fig. \ref{fig:sqann_small}.

\begin{figure}
\centering
\includegraphics[width=0.5\textwidth]{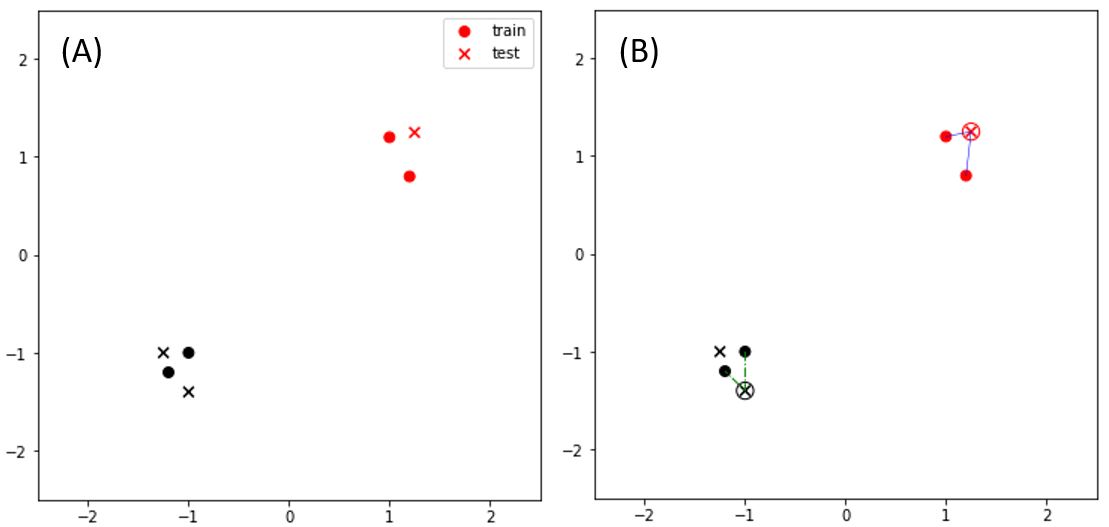}
\caption{(A) Fitting samples are in closed circles, external samples are in x. (B) Open circles show samples that need interpolation, and the corresponding lines point towards the samples from which they are interpolated.} 
\label{fig:sqann_small}
\end{figure}

\subsection{Experiment to test Generalizability of SQANN}
\label{appendix:exptdata}
\textbf{External datasets with increasing spread from fitting dataset distribution}. The accuracy of SQANN on high-dimensional dataset outside the fitting dataset is harder to formalize in theorems. Furthermore, real life data is often noisy and possibly not regularly structured. Hence, for SQANN, we avoid making any statements or any assumptions related to external dataset distributions. 

We instead provide empirical results on external datasets that are \textit{similar} to the fitting dataset, to the extent that each point in the external dataset is a fitting sample perturbed by uniform random values of increasing magnitude. We refer to the noise magnitude as the \textit{test data spread}. Fig. \ref{fig:sqann_expt}(A,A.2) show four domains \(X\) with different test data spread. External dataset that has larger test data spread contains data samples that are noisier and further away from the fitting data points. Fig. \ref{fig:sqann_expt}(B,B.2) show that SQANN naturally performs better with smaller test data spread. As the test data spread increases, larger errors are observed. Likewise, smaller spread means smaller \(N_{interp}\), i.e. fewer data samples fail to activate neurons in SQANN strongly. For all, errors on fitting data samples are 0 \textit{as expected from theorem \ref{theorem:sqann}}.

Remark: a reviewer mentioned that obtaining external samples through the perturbations of fitting samples introduce correlation i.e. samples are not independent. This is not done before as far as we know. We acknowledge that we are not considering random variables in the sense that statisticians usually consider. The rationale behind our choice of external samples is based on a simple observation: an approximation is sometimes only meaningful in the vicinity of known samples. For data samples that are found far from known samples, unless we have a model that pre-defines their relations to existing dataset or a knowledge on the dataset's behaviour, there exists no universal rule that prevent them from taking any values. This is the reason why we do not consider random variables in the ``ordinary" sense; however, note that ``ordinary" is just the popular iid variables statisticians often use. In this paper, we are satisfied with experiments that demonstrate the following: so long as the values around known samples are predicted with reasonable accuracy, we say that our UA models generalize well (on top of accurately remembering existing fitting dataset). Naturally, the stronger the perturbation, the lower accuracy we expect to see.

\textbf{Experimental data specification}. In fig \ref{fig:sqann_expt}(A) [Top row] fitting samples \(X\) are \(x\in \mathbb{R}^2\) with components drawn from uniform r.v. (random variable) \(x_1=x_2=t\sim U(-1.,1.)\) plus Gaussian noise. External samples are \(X\) plus r.v.  from \(U(-s,s)\) where \(s\) is the test data spread. Also, \(y=||x||\) is indicated by the color. [Bottom row] similar to top row, but \(x\in X\) with \(x=R[cos(t),sin(t)]^T\) where \(t\sim U(0, \approx 2\pi),R\sim U(0.8,1.2)\) and \(y=cos(t)\). For both experiments in the top and bottom rows , 128 fitting data samples are used for SQANN construction and tested on 128 external data samples.

\textbf{Classification and visualization of SQANN's special interpretability features}. We show the use of SQANN for a simple classification problem in fig. \ref{fig:sqann}(C). The ring outside is labelled \(0.5\), while the ring inside \(1.0\). With activation parameters \(a_1=a_2=1\), we achieve zero error not only for fitting dataset (to be expected from theorem \ref{theorem:sqann}) but also on external dataset. A special feature in SQANN is its ability to tell the user which data samples fail to activate any neurons strongly; such samples' output must be interpolated (see SQANN propagation). In fig. \ref{fig:sqann}(C) right, points marked with red open circles need interpolation. Each such point is interpolated using two fitting samples whose ``fingerprint" neurons are most strongly excited. These two samples are shown as the two points directly linked by colored straight lines to the open circle. This is possible because SQANN systematically stores indices of fitting data samples within respective layers. The list of indices organized by layers can even be explicitly printed e.g. see SQANN.ipynb, supp. materials.

\begin{table*}[t]
\caption{MSE of different regression methods for Boston Housing dataset. Row o. (original) shows MSE obtained from models trained on \(D\). Row eT shows MSE from models trained on \(D'\) with \(\tau=T\). All are evaluated on \(D_{external}\). Interestingly, ood data identified by SQANN helps improve DTree accuracy significantly.}
\label{table:sqannboston}
\vskip 0.15in
\begin{center}
\begin{small}
\begin{tabular}{l|llllllllll}
\toprule
& Lin & Ridge & Lasso & LSVR & NuSVR &   SVR & DTree & kneigh &    MLP & SQANN \\ 
\midrule
o. &  36.45 & 7.990 & 9.834 &  8.355 & 8.833 & 8.712 &   7.356 &  7.393 & 12.77 & 9.898 \\
e5 &  5.139 & 5.135 & 7.068 &  5.950 & 6.295 & 6.068 &   5.026 &  4.798 &  3.481 & 7.998 \\
e2 &  4.993 & 5.028 & 7.882 &  5.832 & 6.121 & 5.895 &   \textbf{2.072} &  3.025 &  3.846 & \textbf{3.076} \\
\bottomrule
\end{tabular}
\end{small}
\end{center}
\vskip -0.1in
\end{table*}

\textbf{SQANN is tested for regression on Boston Housing and Diabetes Datasets} to demonstrate its generalizability to unseen (external) samples. The procedure is as the following. Let Boston Dataset be \(\mathcal{D}=\{(x^{(k)},y^{(k)}),k=1,\dots,506\}\).  (1) A small set of samples \(D=\{(x^{(k)},y^{(k)}),k=1,...,100\}\) is used to train SQANN and 9 other regression methods; recall that ordered sequence of data is used for SQANN construction.  (2) Mean Squared Errors (MSE) values are measured on unused data \(D_{external}=\mathcal{D}\backslash D\) on all 10 models listed in table \ref{table:sqannboston}. We expect large errors on some external samples, since the fitting dataset might be too small to be representative. (3) SQANN's activations are used to collect samples with large absolute errors \(e_{\tau}(x)=|SQANN(x)-y_0|>\tau\) and we treat them as out-of-distribution (OOD) samples. These samples are considered as new distinct samples to be integrated into \(D\) as the new fitting dataset \(D'\). (4) Train the 10 models, now with \(D'\). (5) Then MSE is measured again on \(D_{external}\) (yes, there will be partial overfitting). Repeat the process for diabetes dataset, where \(|\mathcal{D}|=442\); we also start with \(|D|=100\).

\textbf{Note on table \ref{table:sqannboston}}. The entries in the header denote the models available in scikit-learn (\cite{scikit-learn-apis}): Lin, Ridge and Lasso are the linear models: linear , Ridge (linear least square with L2 regularization),  Lasso (linear with L1) respectively; the Support Vector Regression models: LinSVR, NuSVR, SVR are respectively linear SVR, \(\nu\)-SVR and \(\epsilon\)-SVR. DTree: Decision tree; kneigh: k neighbours are selected from the best \(k=2,3,...,16\), MLP: multi-layer perceptrons, or the fully-connected neural network, with 2 layers, each layer having 64 neurons each trained for a max of 12000 iterations (convergence is attained for both). For SQANN, the initial model trained on \(D\) is kept after SQANN' is trained on \(D'\). Thus, we can choose results based on the strength of activations between SQANN and SQANN'. 

For Boston Housing Dataset \(\tau=5\) (i.e. e5 of table \ref{table:sqannboston}), using SQANN we integrated 211 samples from the external dataset into fitting dataset, so \(|D'|=311\). Overall, \(0.615\) of the whole \(\mathcal{D}\) is used for new training. For \(\tau=2\), i.e. e2 of table \ref{table:sqannboston}, using SQANN we integrated 319 samples, so \(|D'|=419\) i.e. \(0.828\) of the whole \(\mathcal{D}\) is used for new training. With \(\tau=2\), regression performance of SQANN improves greatly compared to other methods, except for decision tree regression. We have thus also seen that SQANN can be used to perform sample selections for data that appear to be out of distribution; this has improved decision tree performance greatly. The performance of other models have improved reasonably too, especially MLP. For MLP, however, the randomness used to achieve convergence to some local minima might have led it to explore other minima; hence we get slightly decreased performance for e2 compared to e5. 

\begin{table}[t]
\begin{center}
\caption{MSE of different regression methods for Diabetes dataset. Notations similar to table \ref{table:sqannboston}.}
\label{table:sqanndiabetes}
\begin{tabular}{l|llllllllll}
    & Lin & Ridge & Lasso & LSVR & NuSVR &   SVR & DTree & kneigh &    MLP & SQANN \\ 
\hline
o.   & 58.57 & 57.60 & 59.21 & 94.65 & 84.34&  81.04  & 76.73 & 66.95 & 108.86 & 93.80 \\
e40  & 54.15 & 54.18 & 55.65 & 70.92 & 74.39 & 73.57  & 40.52 & 53.06 &  54.17 & 53.96 \\
\end{tabular}
\end{center}
\end{table}

For Diabetes Dataset, \(\tau=40\) yields relatively competitive performance. using SQANN we integrated 218 samples, so \(|D'|=318\) so \(0.719\) of the whole \(\mathbb{D}\) is used for new training. From the results of linear models, the dataset appears to have a somewhat strong linear structure; but non-linear models still can perform better.

Update: in version 2 of our repository, we also tested SQANN without variable \(a_1,a_2\) that are used in previous examples (the model is still the same). They result in more absorbed ood samples, but with performance that beats the other methods on both datasets, as shown in table \ref{table:sqann2results}. Results can be found in OANN/examples.

\begin{table}[t]
\begin{center}
\caption{MSE for version 2.  Notations similar to table \ref{table:sqannboston}. B2: Boston, D2: diabetes}
\label{table:sqann2results}
\begin{tabular}{l|ll}
\toprule
B2 & DTree & SQANN \\ 
\midrule
o. &   7.356 &  12.646 \\
e5 &   5.026 &  2.604 \\
e2 &   2.072 &  \textbf{1.270} \\
\hline
D2    & DTree  & SQANN \\ 
\hline
o.   & 76.73 & 93.80 \\
e40  & 40.52 & \textbf{36.159} \\
\end{tabular}
\end{center}
\end{table}

\label{appendix:pseudocode}
\textbf{Regarding the pseudo code}. Mapping names from pseudo code to python code:\newline sample\_collection \(\rightarrow\) layer\_k\_sample\_collection. \newline
push\_node\(\rightarrow\) push\_node\_to\_layer. \newline
forward\_cons \(\rightarrow\) forward\_to\_layer\_k\_for\_reconstruction.\newline
ssig \(\rightarrow\) STOP SIGNAL \newline
collision \(\rightarrow\) COLLISION. \newline
remove\_index \(\rightarrow\) remove\_index\_to\_layer\_node
check\_signal, new\_nodes and update\_nodes are placeholders for simple check and update sequences in python code.

\section{More Comments}
\label{appendix:extra}
Based on ICLR 2022 anonymized reviews, contents that are less relevant are moved here.

\subsection{On Direct generalization of TNN}
\label{extra:TNNgen}

As mentioned in appendix section \ref{tnnremarks}, \textit{our attempts at generalizing TNN directly to high-dimensional input data are unsatisfactory. Readers can skip this without losing any information required for understanding this paper}. For the record, we still list these attempts here. Now, we consider multi-dimensional input \(x\in\mathbb{R}^n, n>1\) with one dimensional output \(y\) for TNN. If the well-known \textit{space-filling curve} can be constructed for the input data, generalization is immediately done. Otherwise, variable separability might help with modelling a system partially using TNN. For example, given an ideal two-variable model \(F(x,y)=f(x)g(y)\). Suppose it fits the experimental data poorly. Assuming that \(f\) is correct, correction can be applied to \(g\) by replacing it with \(g(y)+NN(y)\) to account for the errors. Computing \(TNN(y)=\frac{F(x,y)}{f(x)}-g(y)\) yields the \(y\) values to be used for weight computations in our triangular construction. Clearly, error correction can be performed on \(f\) similarly, though we now have to determine how to allocate values to each factor. 

One way to order multi-dimensional data samples will be ordering by the magnitude of the vector \(x=[x_1,...,x_N]^T\to [r,x_2,...,x_N]^T\) where \(r=\sqrt{\Sigma_i x_i^2}\). The sign of \(x_1\) can be generally dealt with once we include discrete variables (such as binary variables) in the vector, where weights could be toggled to different values according to the discrete variables. More generally, weights that vary with a continuous variable may be a powerful modification to the current method of construction which uses constant weights. This will lead to the loss of linear ordering, in the sense of ``larger" vs ``smaller" in the inequality of real number, leaving us with more relaxed conditions, possibly partial order. However, they will not be within the scope of this paper.

For now, assume we have already defined a linearly ordering for the dataset, we can now generalize the construction. Now \(W\in\mathbb{R}^{N\times n}\), where \(N,n\) are still respectively the number of data samples and dimensions. Similar to the earlier sub-section \textit{computing weights}, \((Wx^{(k+1)}+b)_{N-k+1}=-a\) and \((Wx^{(k)}+b)_{N-k+1}=a\). By subtracting them, we now get \([W(x^{(k)}-x^{(k+1)})]_{N-k+1}=2a\). Spelling it out and rewriting component-wise difference as \(\Delta_i^{(k)}=x^{(k)}-x^{(k+1)}\), we have 
\begin{equation}
\label{multintriangular}
\Sigma_{i=1}^n W_{N-k+1,i}\Delta_i^{(k)}=2a
\end{equation}
Since \(\Delta_i^{(k)}\) can be zero, we only need to carefully choose \(W_{N-k+1,i}\) to satisfy equation (\ref{multintriangular}). This is simply done by setting 
\begin{equation}
W_{N-k+1,i} = \begin{cases} 0 & if\ \Delta_i^{(k)}=0 \\ \frac{2a}{N_1 \Delta_i^{(k)}} & otherwise\end{cases}
\end{equation}
where \(N_1\) is the number of non-zero \(\Delta_i^{(k)}\) over all \(i\) for the particular \(k\). Finally, \(b_{N-k+1}=a-W_{N-k+1} x^{(k)}\) and likewise \(\alpha=A^{-1}y\) in the same manner.

The difficulty is in selecting the choice of the linear ordering. We are not able to provide any sensible ordering for high dimensional dataset, though a meaningful ordering we may still exist. Thus we ask does there exist any linear ordering so that TNN can be used for high accuracy classification? The proposition indicates that there is.

In the following, training/test are interchangeable with fitting/external.

\begin{proposition}. Given a standard DNN for C classes classification with \(a_{tr}\) training accuracy, then there exists a linear ordering for TNN to achieve \(a_{tr}\) accuracy. Test accuracy  \(a_{test}\) of DNN can be achieved by TNN with high probability through \textit{squeezed linear ordering}. \end{proposition}

Proof: In the stringent case, classification is performed by taking \(c=argmax_{i}y_i\) where \(y=DNN(x)\), \(x\) the image to be classified, and \(y\in\mathbb{R}^C\) the output from the last layer of the neural network \(DNN\).  Suppose \(C=10\) and \(c=0,1,\dots,9\). Partition the unit interval \([0,1]\) such that the ten classes are evenly distributed, i.e. \(I_c=[\frac{c}{C},\frac{c+1}{C}]\). For each class \(c\), collect all the training data samples that are classified as \(c\) by the DNN, denote this set as \(P_c\). Then perform the following linear ordering: map \(p_c\) to \(\frac{c+0.5}{C}\) where \(p_c\) is the data point with the highest \(y_c\) component at the last layer of DNN. Let \(R_c=\{r_i\in P_c\backslash\{p_c\}:DNN_{c+1}(r_i)\ge DNN_{c-1}(r_i)\}\). Then, continue constructing the linear ordering \(p_c<r_1<r_2<\dots\) so that \(r_k<r_{k+1}\) whenever \(DNN_c(r_k)\ge DNN_{c}(r_{k+1})\), i.e. going to the right of the interval, a data sample has the less probability of being classified as \(c\) and it is also more likely classified as \(c+1\) than \(c-1\) if \(c\) is excluded. Let \(L_c\) be the remaining set of data samples classified as \(c\), i.e. \(P_c\backslash(R_c\cup\{p_c\})\). Set \(\dots <l_2<l_1<p_c\) so that \(l_{k+1}<l_k\) whenever \(DNN_c(l_k)\ge DNN_{c}(l_{k+1})\). The linear ordering for training dataset is done and TNN will yield \(a_{tr}\). 

\textit{Squeezed linear ordering}. We can map the ordered samples to the interval \(I_c\) arbitrarily, so long as the order is preserved. However, placing each \(R_c,L_c\) nearer to \(p_c\) increases TNN chances of achieving \(a_{test}\), i.e. we squeeze the linear ordering towards the centers \(p_c\) for all \(c\). This is because, ordering external data samples into the linear order made by training samples, we may get the situation where external sample has max \(y_c\) (thus likely classified as \(c\)), but lies beyond the extreme end of \(R_c\) or \(L_c\). Conversely, such sample is beyond the extreme end of \(L_{c+1}\) or \(R_{c-1}\) respectively. By squeezing, large spaces (the \textit{grey area}) are reserved between the extreme ends of two classes. Data samples could fall into these intervals, and we reduce the possibility of misclassifying edge cases, e.g. by letting users treat them separately when TNN indicates that these samples lie in the \textit{grey area}. \(\square\)

Since TNN can achieve arbitrary accuracy on the training dataset, we will recover the same accuracy as the DNN from the newly ordered training data. We do not quantify the exact probability of attaining high \(a_{test}\) because distribution of external dataset may not be exactly the same as the training dataset. With DNN as the encoder mapping the raw data into the unit interval is ironic since the hardwork is already spent on training the deep neural network. However, this shows that, in principle, there exist encodings that can achieve linear ordering for TNN to attain high accuracy classifications. Notes: when clashes occur, e.g. \(r_k\) and \(r_{k+1}\) have the same \(DNN_c\) components, for the purpose of this simple construction, simply randomly assign order between them. We only aim to prove the existence of linear ordering here.

\end{document}